\documentclass[lettersize,journal]{IEEEtran}
\usepackage{amsmath,amsfonts}
\usepackage{cite}
\usepackage{array}
\usepackage[caption=false,font=normalsize,labelfont=sf,textfont=sf]{subfig}
\usepackage{textcomp}
\usepackage{stfloats}
\usepackage{url}
\usepackage{verbatim}
\usepackage{graphicx}
\usepackage{algorithm}
\usepackage{algpseudocode}
\usepackage{xcolor}
\def\BibTeX{{\rm B\kern-.05em{\sc i\kern-.025em b}\kern-.08em
    T\kern-.1667em\lower.7ex\hbox{E}\kern-.125emX}}
\usepackage{balance}

\usepackage{xcolor}

\DeclareMathAlphabet\mathbfcal{OMS}{cmsy}{b}{n}

\newcommand{\mat}[1]{\mathbf{#1}}
\usepackage[final,commandnameprefix=ifneeded]{changes}
\usepackage{xcolor}

\definechangesauthor[name={Author}, color=blue]{auth}
\definechangesauthor[name={R1}, color=red]{r1}
\definechangesauthor[name={R2}, color=green]{r2}

\begin{document}
\title{DeepOHeat-v1: Efficient Operator Learning for\\ Fast and Trustworthy Thermal Simulation and Optimization in 3D-IC Design}
\author{Xinling Yu, Ziyue Liu, Hai Li, Yixing Li, Xin Ai, Zhiyu Zeng, Ian Young \IEEEmembership{Fellow, IEEE}, Zheng Zhang
\thanks{This work is sponsored by Technology Research (TR) and Business Development Strategic Research Sector (BD SRS) of Intel Corporation.}
\thanks{Xinling Yu and Zheng Zhang are with the Department of Electrical and Computer Engineering, University of California, Santa Barbara, CA 93106 USA (email: xyu644@ucsb.edu; zhengzhang@ece.ucsb.edu).}
\thanks{Ziyue Liu is with the Department of Computer Science, University of California, Santa Barbara, CA 93106 USA (email: ziyueliu@ucsb.edu).}
\thanks{Yixing Li, Xin Ai, and Zhiyu Zeng are with Cadence Design Systems, Austin, TX 78759 USA (email: yixingli@cadence.com; nathanai@cadence.com; zzeng@cadence.com).} 
\thanks{Hai Li and Ian Young are with Intel Corporation, Hillsboro, OR
97124 USA (email: hai.li@intel.com; ian.young@intel.com).}
}

\maketitle

\begin{abstract}
Thermal analysis is crucial in three-dimensional integrated circuit (3D-IC) design due to increased power density and complex heat dissipation paths. Although operator learning frameworks such as DeepOHeat~\cite{liu2023deepoheat} have demonstrated promising preliminary results in accelerating thermal simulation, they face critical limitations in prediction capability for multi-scale thermal patterns, training efficiency, and trustworthiness of results during design optimization. This paper presents DeepOHeat-v1, an enhanced physics-informed operator learning framework that addresses these challenges through three key innovations. First, we integrate Kolmogorov-Arnold Networks with learnable activation functions as trunk networks, enabling an adaptive representation of multi-scale thermal patterns. This approach achieves a $1.25\times$ and $6.29\times$ reduction in error in two representative test cases. Second, we introduce a separable training method that decomposes the basis function along the coordinate axes, achieving $62\times$ training speedup and $31\times$ GPU memory reduction in our baseline case, and enabling thermal analysis at resolutions previously infeasible due to GPU memory constraints. Third, we propose a confidence score to evaluate the trustworthiness of the predicted results, and further develop a hybrid optimization workflow that combines operator learning with finite difference (FD) using Generalized Minimal Residual (GMRES) method for incremental solution refinement, enabling efficient and trustworthy thermal optimization. Experimental results demonstrate that DeepOHeat-v1 achieves accuracy comparable to optimization using high-fidelity finite difference solvers, while speeding up the entire optimization process by $70.6\times$ in our test cases, effectively minimizing the peak temperature through optimal placement of heat-generating components. Open source code is available at \url{https://github.com/xlyu0127/DeepOHeat-v1}.
\end{abstract}


\section{Introduction}
\IEEEPARstart{T}{hree-dimensional} integrated circuit (3D-IC) is a promising solution to address the scaling challenges of semiconductor chips\cite{3dic_benefits}. Although vertical stacking of multiple active silicon layers enables a higher integration density and improved system performance \cite{3dic_review}, it also introduces critical thermal management challenges \cite{optimal, survey,survey2}. Heat generated in one layer must pass through multiple layers before reaching the heat sink, creating thermal bottlenecks and potential reliability issues \cite{3dic_thermal_reliability}. Addressing these thermal challenges requires extensive analysis and optimization during the design phase. This process involves solving partial differential equations for numerous design configurations, making it computationally intensive. Traditional numerical methods, such as finite element method and finite difference method, while accurate, are too time-consuming for iterative design optimization \cite{survey2, liu2014compact, li2004efficient}. This has motivated the development of various surrogate models, including model order reduction \cite{wang2004spice,xie2013system} and data-driven approaches \cite{dnn-thermal, ml-based-3d, ml-thermal}. \added{Recent advances in multiscale thermal modeling, such as the ROM-based approach by \cite{geb2022chip}, have demonstrated the importance of capturing disparate length scales in 3D-IC thermal analysis.}

Neural network-based methods can perform real-time predictions for thermal analysis, but face significant limitations. Data-driven approaches \cite{wangaro, smith2023real, ml-thermal} require extensive simulation datasets and lack generalization. Physics-informed neural network-based methods \cite{unsupervised-auto-encoder} reduce data dependency but are limited to specific geometric parameters. Most existing works either fail to solve 3D full-chip temperature fields or cannot handle different boundary conditions and power distributions. Autoencoder-decoder methods \cite{smith2023real, unsupervised-auto-encoder} work for 2D prediction but fail with 3D volumetric power maps, while graph neural networks \cite{sanchis2022towards} still require substantial labeled data despite better spatial modeling.

Recently, operator learning frameworks like DeepOHeat \cite{liu2023deepoheat} and its variant \cite{lu2024fast} have shown promise in addressing these limitations. DeepOHeat learns a mapping from design configurations to temperature fields through physics-informed training, eliminating the need for expensive simulation data and effectively handling 3D geometries. Despite these advances, DeepOHeat faces three critical limitations. First, the multilayer perceptron (MLP) trunk networks exhibit inherent spectral bias \cite{ffn, wang2021eigenvector}, limiting their ability to capture multi-scale thermal patterns that are crucial for accurate temperature prediction. Second, its physics-informed training requires computing high-order derivatives across numerous collocation points, making it memory-intensive and time-consuming. This computational burden prevents its application to large-scale problems with high-resolution thermal analysis. Third, like most operator learning frameworks, DeepOHeat lacks mechanisms to assess prediction trustworthiness during optimization, potentially leading to suboptimal or unreliable design decisions. 

This paper presents \emph{DeepOHeat-v1} to address the above limitations. Through architectural and algorithmic innovations, DeepOHeat-v1 can enhance both computational efficiency and prediction reliability of operator learning for thermal analysis, and enable thermal optimization for 3D-IC floorplanning with both low computational cost and high accuracy. Our key contributions are summarized below.
\begin{itemize}
\item Accuracy improvement through Kolmogorov-Arnold networks (KAN). Unlike MLPs with fixed activation functions, KANs employ learnable univariate functions\cite{liu2024kan,ss2024chebyshev}, allowing a more adaptive representation of multiscale thermal patterns. We express KAN as DeepOHeat trunk networks, achieving  $1.25\times$ and $6.29\times$ error reductions in two representative test cases.
\item Efficiency improvement via separable training. We employ recently developed separable training method \cite{yu2024separable} to decompose the basis functions of DeepOHeat along coordinate axes. Using forward-mode automatic differentiation \cite{khan2015vector}, this approach achieves $62\times$ training speedup and $31\times$ memory reduction in our baseline case, making high-resolution thermal analysis computationally feasible.
\item Trustworthy thermal optimization. Lack of trustworthiness is a fundamental challenge that prevents the application of almost all neural network-based surrogates in design optimization. We propose a confidence estimator to evaluate whether a DeepOHeat prediction is acceptable or not at every step of an optimization loop. Once a prediction is not acceptable (for some corner design cases), we use it as an initial guess and call GMRES (the Generalized Minimal Residual Method~\cite{saad1986gmres}) for only a small number of iterations to get a high-fidelity finite-difference (FD) solution. This hybrid framework achieves an excellent trade-off between computational efficiency and solution accuracy. We apply this optimization framework to \replaced{optimize thermal characteristics in 3D chip designs}{optimize the floorplan of 3D-IC for improved thermal characteristics} and obtain an accuracy comparable to that of an FD-based optimizer while achieving $70.6\times$ speedup.
\end{itemize}

\section{Background, Related Work, and Motivation for DeepOHeat-v1}
This section first presents the fundamental PDE governing thermal behavior in 3D-ICs, then reviews how DeepOHeat \cite{liu2023deepoheat} reformulates this problem through operator learning, highlighting its strengths and limitations. These limitations motivate our proposed enhancements in DeepOHeat-v1.

\subsection{Thermal Simulation in 3D-ICs}
Thermal simulation plays a fundamental role in 3D-IC design, where accurate temperature prediction is essential for reliability and performance optimization. The temperature distribution within a chip domain $S$ is governed by the heat conduction equation:
\begin{equation}
\frac{\partial}{\partial y_1}\left(k \frac{\partial T}{\partial y_1}\right) + \frac{\partial}{\partial y_2}\left(k \frac{\partial T}{\partial y_2}\right) + \frac{\partial}{\partial y_3}\left(k \frac{\partial T}{\partial y_3}\right) + q_V = \rho c_p \frac{\partial T}{\partial t},
\label{heat:general}
\end{equation}  
where $T$ represents the temperature field and $q_V = q_V(\mat{y}, t)$ denotes the volumetric heat generation rate at any spatial-temporal coordinate $(y_1, y_2, y_3, t) \equiv (\mat{y}, t)$. The material-specific properties $k$, $\rho$, and $c_p$ correspond to thermal conductivity, mass density, and heat capacity, respectively.

In this work, we consider heterogeneous multi-layer chip architectures where the thermal conductivity varies vertically across different material layers. Mathematically, this variation is captured by expressing $k$ as a piecewise function of the vertical coordinate: $k = k(y_3)$. Within each layer, we assume isotropic thermal conductivity, meaning that heat conduction is uniform in all directions, i.e. $k_{y_1} = k_{y_2} = k_{y_3} = k(y_3)$.

\added{In this work, we focus on steady-state thermal analysis.} Under steady-state conditions ($\frac{\partial T}{\partial t} = 0$), Equation \eqref{heat:general} reduces to:
\begin{equation}
    \nabla \cdot (k(y_3) \nabla T) + q_V(\mat{y}) = 0,
    \label{heat}
\end{equation}  
where $\nabla \cdot (k(y_3) \nabla T)$ accounts for the variation of thermal conductivity across layers. 

The thermal behavior at the chip boundaries is governed by appropriate boundary conditions (BCs) that reflect the thermal management strategy. In this work, we consider three fundamental types of BCs:
\begin{itemize}
\item \textbf{Neumann BC}, specifying heat flux at the surface:
\begin{equation}
-k(y_3)\frac{\partial T}{\partial y_i} = q_n,
\label{eq:neumann}
\end{equation}
where $q_n$ represents the local heat flux density. This condition is particularly important for modeling surface power maps in 3D-ICs.
\item \textbf{Adiabatic BC}, representing perfectly insulated surfaces:
\begin{equation}
\frac{\partial T}{\partial y_i} = 0.
\label{eq:adiabatic}
\end{equation}
\item \textbf{Convection BC}, modeling heat exchange with the ambient environment:
\begin{equation}
-k(y_3)\cdot \frac{\partial T}{\partial y_i} = h(T-T_{\rm amb}),
\label{eq:convection}
\end{equation}
where $h$ denotes the heat transfer coefficient and $T_{\rm amb}$ represents the ambient temperature.

\end{itemize}

\subsection{DeepOHeat for Thermal Analysis}
The thermal system described above presents significant computational challenges for design optimization. Although traditional mesh-based numerical methods provide accurate solutions, they are computationally expensive, particularly for iterative design evaluations. To address this challenge, DeepOHeat \cite{liu2023deepoheat} reformulates thermal simulation through operator learning, enabling rapid temperature prediction without requiring expensive simulation data.

\subsubsection{Thermal Simulation as an Operator Learning Problem}
\begin{figure}[t]
    \centering
    \includegraphics[width=\linewidth]{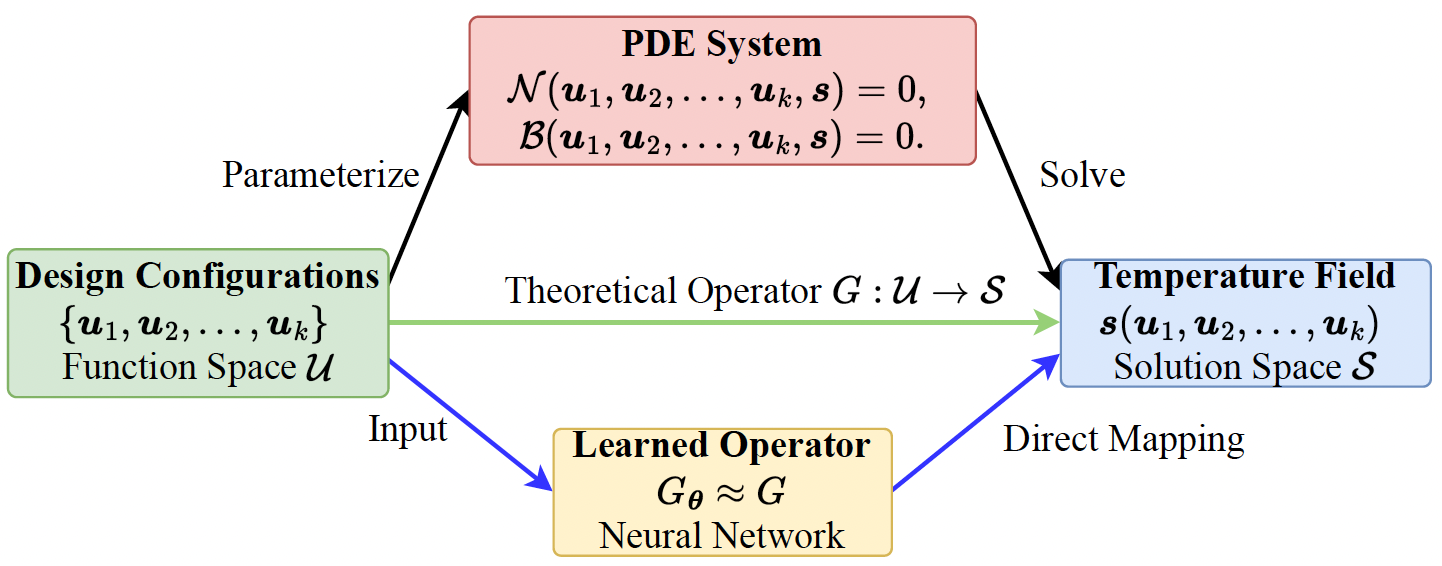}
    \caption{Operator learning framework for thermal simulation. The theoretical operator $G$ provides a direct mapping from design configurations to temperature fields and is fundamentally defined by the PDE system. The objective of operator learning is to train a surrogate neural operator $G_{\boldsymbol{\theta}}$ that approximates $G$, enabling rapid temperature prediction without solving PDEs for each new design configuration.}
    \label{fig:operator_learning}
\end{figure}

As illustrated in Fig.  \ref{fig:operator_learning}, the thermal system governed by Equations \eqref{heat}-\eqref{eq:convection} can be be formulated in operator form:
\begin{equation}
\begin{aligned}
\mathcal{N}(\boldsymbol{u}_1, \boldsymbol{u}_2, \ldots, \boldsymbol{u}_k, \boldsymbol{s}) = 0,\\
~\mathcal{B}(\boldsymbol{u}_1, \boldsymbol{u}_2, \ldots, \boldsymbol{u}_k, \boldsymbol{s}) = 0,
\end{aligned}
\label{eq:pde_system}
\end{equation}
where $\mathcal{N}$ represents the heat equation operator, $\mathcal{B}$ denotes the boundary operators, and $\boldsymbol{s}$ is the temperature field. The system is parameterized by $k$ design configurations $\{\boldsymbol{u}_i\}_{i=1}^k$, which can be either parametric (e.g., heat transfer coefficients) or non-parametric (e.g., power distributions).

In the traditional approach, solving this PDE system requires expensive numerical computations for each new set of design configurations, as shown in the upper path of Fig.~\ref{fig:operator_learning}. Given these PDE configurations $\{\boldsymbol{u}_i\}_{i=1}^k$, the temperature field evaluated at spatial coordinates $\mat{y}$ is denoted as:
\begin{equation}
    \boldsymbol{s} (\boldsymbol{u}_1, \boldsymbol{u}_2, \ldots, \boldsymbol{u}_k)(\mat{y}).
\end{equation}
The key assumption of the operator learning framework is that there exists a unique operator $G$ that maps the function space $\mathcal{U} = \mathcal{U}_1 \times \mathcal{U}_2 \times \dots \times \mathcal{U}_k$, spanned by the PDE configurations, to the solution space $\mathcal{S}$, such that:
\begin{equation}
G: \mathcal{U} \rightarrow \mathcal{S},
\label{eq:operator}
\end{equation}
where $\boldsymbol{s}(\boldsymbol{u}_1, \boldsymbol{u}_2, \ldots, \boldsymbol{u}_k) = G(\boldsymbol{u}_1, \boldsymbol{u}_2, \ldots, \boldsymbol{u}_k)$. The objective of operator learning is to train a surrogate operator model $G_{\boldsymbol{\theta}}$ that approximates $G$. Once trained, $G_{\boldsymbol{\theta}}$ can instantly predict the temperature distribution for new design configurations, significantly accelerating thermal analysis.

\subsubsection{DeepOHeat Framework}
\begin{figure*}[t]
    \centering
    \includegraphics[width=\linewidth]{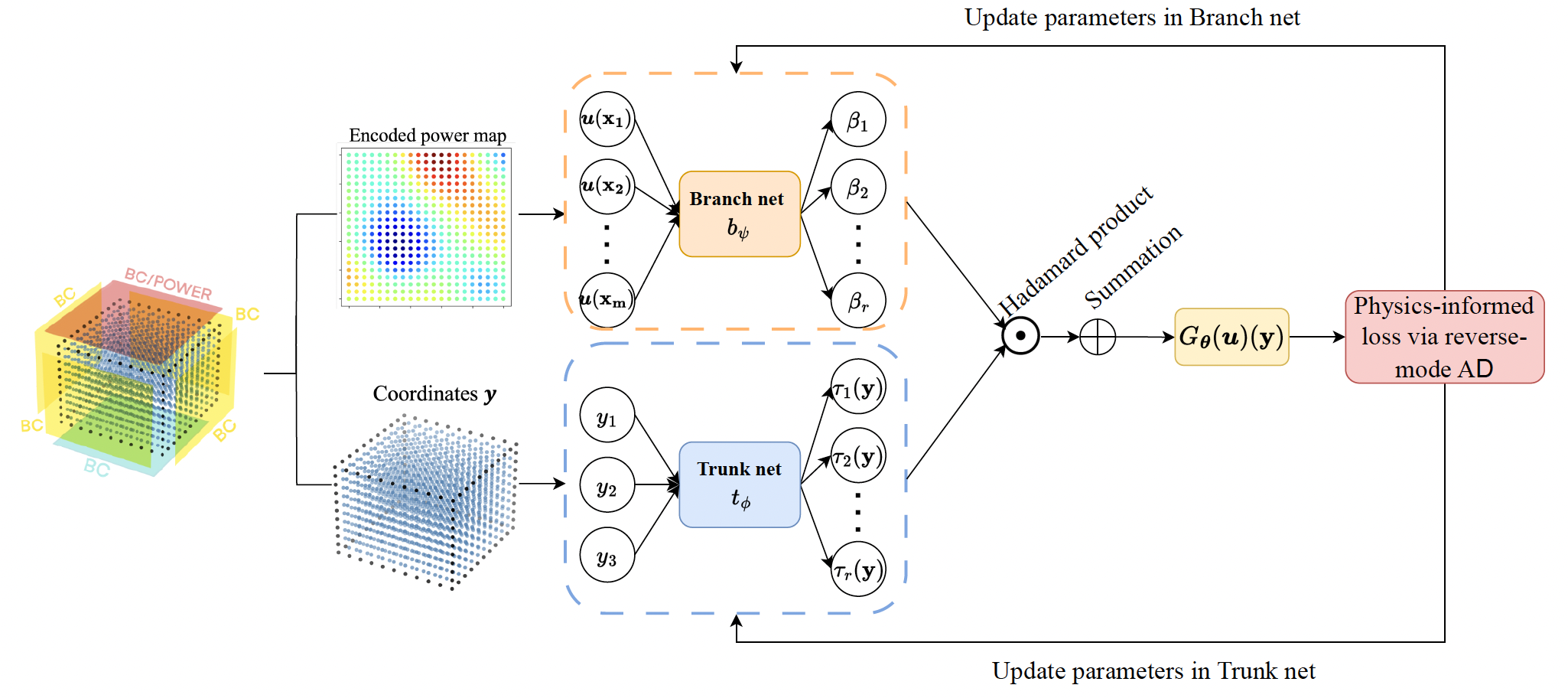}
    \caption{DeepOHeat for thermal simulation. This example shows the case where $k=1$, with the surface power map as the design configuration. The architecture consists of branch and trunk networks that process the encoded power map $\mathcal{E}(\boldsymbol{u})$ and spatial coordinates respectively. Their outputs are combined through Hadamard product and summation to predict temperature, while physics-informed training with reverse-mode AD enables parameter updates without simulation data.}
    \label{fig:deepoheat_arch}
\end{figure*}

DeepOHeat employs a operator learning architecture based on DeepONet \cite{deeponet,jin2022mionet} to construct the temperature field approximation, as illustrated in Fig.~\ref{fig:deepoheat_arch}. The framework combines $k$ branch networks that processes $k$ design configurations with a trunk network that handles spatial dependencies, enabling efficient learning of the thermal operator mapping. Without loss of generality, we consider the case $k=1$ to illustrate the core methodology.

As shown in the figure, the temperature field approximation process consists of three key components. First, an encoder $\mathcal{E}:\mathcal{U}\rightarrow \mathbb{R}^m$ maps the input function $\boldsymbol{u}$ (e.g., surface power) to its point-wise evaluations at $m$ fixed sensor locations $\mat{x_{1}}, \mat{x_{2}}, \ldots, \mat{x_{m}}$ (e.g., some grid points of a surface), yielding $(\boldsymbol{u}(\mat{x_1}),\boldsymbol{u}(\mat{x_{2}}),\ldots,\boldsymbol{u}(\mat{x_m}))=\mathcal{E}(\boldsymbol{u})$. This encoded representation captures the essential features of the design.

This encoded representation is then processed by two specialized neural networks. The branch network $b_{\psi}:\mathbb{R}^m\rightarrow\mathbb{R}^r$, parameterized by $\psi$, transforms the encoded input into $r$ feature coefficients $\left(\beta_{1}, \beta_{2}, \ldots, \beta_{r}\right)$. Simultaneously, the trunk network $t_\phi:\mathbb{R}^d\rightarrow \mathbb{R}^r$, parameterized by $\phi$, maps spatial coordinates $\mat{y}=(y_1, y_2, y_3)$ to $r$ basis functions $\left(\tau_{1}(\mat{y}), \tau_{2}(\mat{y}), \ldots, \tau_{r}(\mat{y})\right)$. The predicted temperature field is then computed as:
\begin{equation}\label{eq:deeponet_prediction}
G_{\boldsymbol{\theta}}(\boldsymbol{u})(\mat{y}) = \sum_{k=1}^r \beta_k \tau_k = b_{\psi}(\mathcal{E}(\boldsymbol{u}))\cdot t_{\phi}(\mat{y}),
\end{equation}
where $\boldsymbol{\theta}=(\psi,\phi)$ encompasses all trainable parameters and $\cdot$ denotes vector dot product.

A key advantage of DeepOHeat, as illustrated on the right side of Fig. \ref{fig:deepoheat_arch}, is its physics-informed training strategy \cite{raissi2019physics,wang2021learning} that eliminates the need for expensive simulation data. Given a set of collocation points $\mathcal{D}$ comprising $N_f$ input designs $\{\boldsymbol{u}^{(i)}\}_{i=1}^{N_f}$ sampled from $\mathcal{U}$, $N_r$ residual points $\{\mat{y}_r^{(j)}\}_{j=1}^{N_r}$ within the chip domain, and $N_b$ boundary points $\{\mat{y}_b^{(j)}\}_{j=1}^{N_b}$, the model parameters are optimized by minimizing the physics-informed loss:

\begin{equation}
L(\boldsymbol{\theta} | \mathcal{D}) = L_{r}(\boldsymbol{\theta} | \mathcal{D}) + L_{b}(\boldsymbol{\theta} | \mathcal{D}),
\label{eq:total_physics_loss}
\end{equation}
where the residual and boundary losses are defined as:
\begin{equation}
\begin{aligned}
L_{r}(\boldsymbol{\theta} |\mathcal{D}) &= \frac{1}{N_{f} N_{r}}\sum_{i=1}^{N_f} \sum_{j=1}^{N_r} \Big| \mathcal{N} (\boldsymbol{u}^{(i)},G_{\boldsymbol{\theta}}(u^{(i)})(\mat{y}_{r}^{(j)})) \Big|^{2},\\
L_{b}(\boldsymbol{\theta} | \mathcal{D}) &= \frac{1}{N_{f} N_{b}}\sum_{i=1}^{N_f} \sum_{j=1}^{N_{b}} \Big| \mathcal{B} (\boldsymbol{u}^{(i)},G_{\boldsymbol{\theta}}(u^{(i)})(\mat{y}_{b}^{(j)})) \Big|^{2}.
\end{aligned}
\label{eq:physics_loss}
\end{equation}

\subsection{Challenges in DeepOHeat for 3D-IC Thermal Analysis} Despite its success in accelerating thermal simulation, DeepOHeat faces three critical limitations that hinder its practical application in thermal-aware design optimization.

First, the accuracy of DeepOHeat heavily depends on the quality of basis functions $\{\tau_k(\mat{y})\}_{k=1}^r$ produced by its trunk network $t_\phi$. However, the multilayer perceptron (MLP) architecture used in the trunk network exhibits inherent spectral bias \cite{ffn,wang2021eigenvector}, preferentially learning low-frequency components before high-frequency ones \cite{ronen2019convergence}. This limitation significantly impacts DeepOHeat's ability to capture complex thermal patterns with varying spatial frequencies, particularly in regions with sharp temperature gradients or localized hotspots.

Second, although DeepOHeat eliminates the need for simulation data through physics-informed training, this approach introduces significant computational challenges. The physics loss optimization requires computing second-order derivatives with respect to numerous collocation points through reverse-mode automatic differentiation (AD) \cite{baydin2018automatic}. For a typical 3D thermal simulation discretized on a $101 \times 101 \times 101$ spatial grid, backpropagating through the computational graph across $N_f$ chip configurations and millions of collocation points becomes prohibitively expensive. This computational burden leads to long training time and excessive memory cost, preventing DeepOHeat's application to high-resolution thermal analysis.

Finally, while DeepOHeat enables rapid temperature field prediction through \eqref{eq:deeponet_prediction}, it is unclear whether an individual prediction is accurate enough or not. This limitation becomes critical in thermally aware design optimization, where inaccurate predictions could lead to suboptimal or potentially unsafe designs. Without a systematic approach to quantify prediction trustworthiness and refine or correct potentially inaccurate results, the practical deployment of DeepOHeat in automated design workflows remains challenging.

To overcome these fundamental constraints, we propose \emph{DeepOHeat-v1}, which introduces three key enhancements:
\begin{itemize}
    \item \textbf{Kolmogorov–Arnold Networks (KAN)} to mitigate the trunk network’s spectral bias and accurately capture multi-scale thermal patterns.
    \item A \textbf{separable training} strategy to drastically reduce computational and memory overhead in physics-informed training for high-resolution 3D-IC analysis.
    \item A \textbf{hybrid optimization workflow} that integrates the operator learning with GMRES-based refinement to ensure both efficiency and trustworthiness during design optimization.
\end{itemize}
We detail these enhancements in the following sections.

\section{Kolmogorov–Arnold Networks with Multi-Frequency Basis Functions}
\label{subsec:kan}
A major challenge in learning thermal fields is to capture both smooth varying regions and sharply localized hotspots. Multilayer perceptrons (MLPs) often struggle with such multi-scale patterns due to spectral bias \cite{ffn}, preferring low-frequency solutions over high-frequency ones \cite{ronen2019convergence}. \added{In this context, 'multi-scale' refers to thermal fields that simultaneously contain low-frequency spatial components (smooth, gradually-varying regions) and high-frequency components (sharp temperature gradients and localized hotspots). The challenge lies not in having features of vastly different sizes, but in accurately representing both the smooth and sharp frequency characteristics that coexist in thermal solutions, particularly around heat sources and material interfaces where sharp gradients naturally occur even with similar-magnitude heating.} To address this, DeepOHeat-v1 uses a Kolmogorov–Arnold Network (KAN) \cite{liu2024kan} as a trunk network. Kolmogorov's representation theorem \cite{schmidt2021kolmogorov} states that any continuous multivariate function $f: \mathbb{R}^n \rightarrow \mathbb{R}$ can be represented as:
\begin{equation}
f(\mat{y}) = \sum_{j=1}^{2n+1} g_j\left(\sum_{i=1}^n \phi_{i,j}(y_i)\right),
\label{eq:kolmogorov}
\end{equation}
where $g_j$ and $\phi_{i,j}$ are continuous univariate functions, and $\mat{y} = (y_1,\ldots,y_n) \in \mathbb{R}^n$. This representation suggests constructing complex multivariate functions through compositions of univariate functions.

\subsection{KAN Implementation for Thermal Analysis}
\begin{figure}[t]
    \centering
    \includegraphics[width=\linewidth]{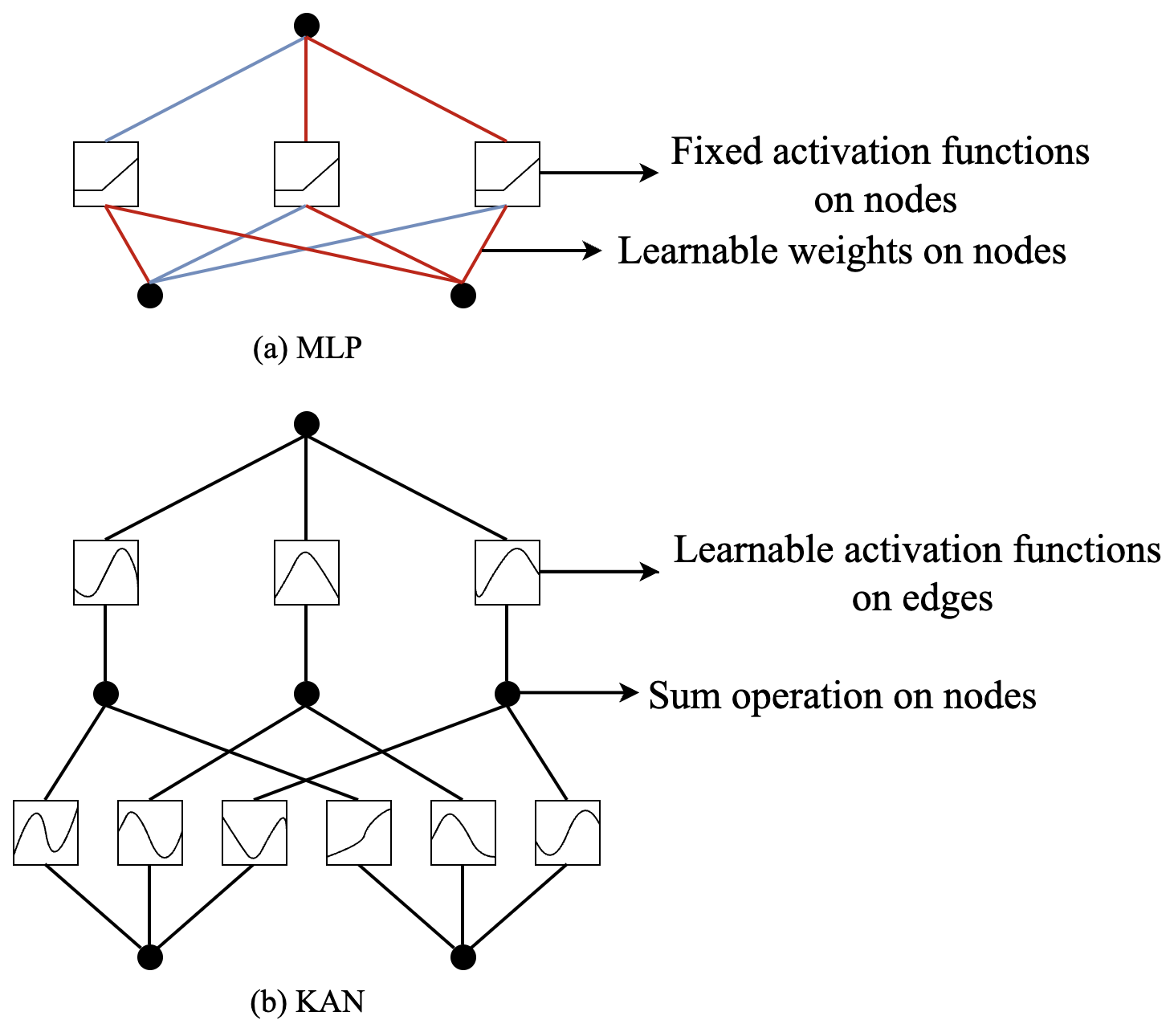}
    \caption{Comparison between (a) a traditional multilayer perceptron (MLP) and (b) a Kolmogorov–Arnold Network (KAN). In MLPs, the activation functions are fixed at the nodes, while weights are learnable. In contrast, KANs use learnable activation functions on edges and sum operations at nodes, offering greater flexibility in function representation.}
    \label{fig:mlp_kan}
\end{figure}

In DeepOHeat-v1, we implement a KAN-based trunk network that directly maps spatial coordinates to basis functions $\{\tau_k\}_{k=1}^r$. In Fig. \ref{fig:mlp_kan}, our KAN implementation differs fundamentally from standard MLPs by using learnable univariate functions on the edges rather than fixed activation functions with trainable weights. Each KAN layer transforms inputs as:
\begin{equation}
y_{j}^{(\ell+1)}
=
\sum_{i=1}^{n_\ell}
\phi_{\ell,i,j}\left(y_i^{(\ell)}\right),
\label{eq:KAN_edge}
\end{equation}
where each trainable function $\phi_{\ell,i,j}$ operates on a single variable $y_i^{(\ell)}$ and is parameterized using an order-$K$ Chebyshev polynomial series:
\begin{equation}
\label{eq:phi_chebyshev}
\begin{aligned}
\phi_{\ell,i,j}(y) &= \sum_{k=0}^{K} a_{\ell,i,j,k}C_k(y), \\
C_k(y) &= \cos(k\arccos(y)), \quad y \in [-1,1],
\end{aligned}
\end{equation}
where $a_{\ell,i,j,k} \in \mathbb{R}$ are trainable coefficients. This structure inherently supports multi-scale representation through its Chebyshev polynomial basis, enabling efficient modeling of both global thermal patterns and localized hotspots. The architecture's ability to naturally represent different frequency scales eliminates the need for explicit Fourier feature mapping \cite{ffn} or deep architectures typically required in MLPs.

\subsection{Theoretical Understanding of the Training Dynamics}
To understand why our KAN implementation effectively captures multi-scale thermal patterns, consider a function with a Chebyshev series expansion:
\begin{equation}
    \label{eq:target_function}
    f^*(y) = \sum_{k=0}^{\infty} b_k C_k(y),
\end{equation}
where $\{b_k\}$ are the true spectral coefficients. To understand the learning dynamics of KANs with Chebyshev basis functions, consider a simplified single-layer model:
\begin{equation}
    \label{eq:single_layer_KAN}
    f(y) = \sum_{k=0}^{K} a_k\, C_k(y), \quad y \in [-1,1],
\end{equation}
where $a_k$'s are trainable parameters and $C_k(y)$'s are the Chebyshev polynomials defined in Eq.~\eqref{eq:phi_chebyshev}. 
 Our network approximates the first $K\!+\!1$ components of $f^*$ through gradient descent. Understanding this multi-frequency learning process provides insights into the behavior of deeper KAN architectures.

The Neural Tangent Kernel (NTK) characterizes how network outputs change with respect to parameters during training \cite{jacot2018neural,lee2019wide}. For our model, the NTK is:
\begin{equation}
    \label{eq:NTK_def}
    \Theta_{\text{KAN}}(y,y') = \sum_{k=0}^{K} \frac{\partial f(y)}{\partial a_k}\,\frac{\partial f(y')}{\partial a_k} = \sum_{k=0}^{K} C_k(y) C_k(y').
\end{equation}
Under gradient flow (see Appendix \ref{app:training_dynamics}), the output evolves as:
\begin{equation}
\label{eq:grad_flow}
\frac{\partial f(y,t)}{\partial t} = -\eta \int_{-1}^{1} \Theta_{\text{KAN}}(y,y')\left(f(y',t)-f^*(y')\right)\frac{dy'}{\sqrt{1-y'^2}},
\end{equation}
where $t$ is the training time and $\eta > 0$ is the learning rate. The error $e(y,t) = f(y,t)-f^*(y)$ decays spectrally (see Appendix \ref{app:training_dynamics}):
\begin{equation}
    \label{eq:error_decomp}
    e(y,t) = \sum_{k=0}^{K} e_k(0)\, e^{-\eta\kappa_k t}\, C_k(y),
\end{equation}
where $e_k(0)$ is the projection of initial error onto the Chebyshev basis, $\kappa_0 = \pi$, and $\kappa_k = \pi/2$ for $k \geq 1$. Unlike MLPs with fixed activation functions, where high-frequency components typically have exponentially smaller eigenvalues \cite{ronen2019convergence}, KANs maintain uniform eigenvalues $\eta\kappa_k = \eta\pi/2$ across all non-constant modes ($k \geq 1$). This enables simultaneous learning of both low- and high-frequency features, making them particularly well-suited for thermal field prediction.

\section{Scaling up DeepOHeat via Separable Training}
Physics-informed training in DeepOHeat requires computing high-order spatial derivatives to evaluate residual and boundary terms in the loss function in Eq. \eqref{eq:physics_loss}. This necessitates automatic differentiation (AD), which applies the chain rule to compute exact derivatives through the neural network's computational graph, providing mathematically precise gradients essential for enforcing differential equation constraints.

However, applying AD in DeepOHeat presents significant computational challenges. For thermal simulation in 3D-ICs, the common-used reverse-mode AD typically employed in neural networks becomes prohibitively expensive when computing second-order derivatives across numerous collocation points. This results in excessive memory consumption and computational overhead that severely limits DeepOHeat's applicability to high-resolution thermal analysis.

To address the  challenges in physics-informed training in DeepOHeat, we propose to decompose the spatial representation along the coordinate axes. This leads to a more efficient separable operator network \cite{yu2024separable}, where the basis functions are learned independently for each spatial dimension.

\subsection{Separable Representation for Single Points}
For a given spatial coordinate $\mat{y}=(y_1,y_2,y_3)$ in a 3D-IC design, we approximate the temperature with a separable representation:
\begin{equation}
\begin{aligned}
\label{eq:scalar_sep}
&G_{\boldsymbol{\theta}}(\boldsymbol{u})(\mat{y}) = \sum_{k=1}^{r} \beta_k \prod_{n=1}^3 \tau_{n,k}\\
&=b_\psi(\mathcal{E}(\boldsymbol{u}))\cdot \left(t^1_{\phi_1}(y_1)\odot t^2_{\phi_2}(y_2)\odot t^3_{\phi_3}(y_3)\right).
\end{aligned}
\end{equation}
Here $\odot$ denotes the Hadamard product\footnote{The Hadamard product denotes the element-wise multiplication of vectors or matrices of the same dimensions.}.  Unlike DeepOHeat that uses a single trunk network, we employ three independent networks $t^n_{\phi_n}:\mathbb{R}\rightarrow\mathbb{R}^r$, each dedicated to learning basis functions $\tau_{n,k}$ for one spatial dimension. This separation allows for more efficient learning of spatial dependencies and enables parallel processing of spatial dimensions.

\subsection{Grid-based Evaluation}
\begin{figure*}[t]
    \centering
\includegraphics[width=0.7\linewidth]{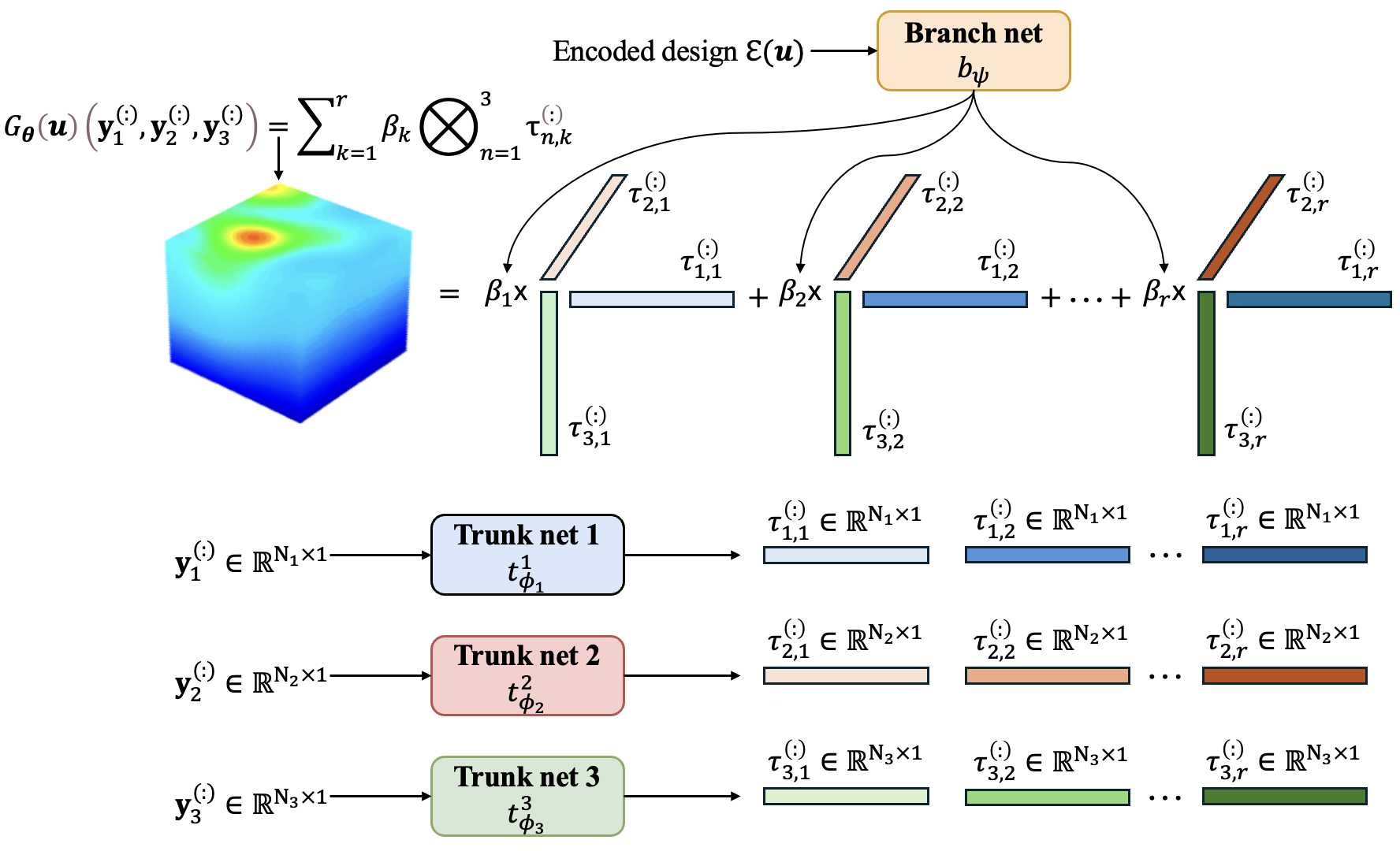}
    \caption{Separable representation for thermal prediction in 3D-ICs. The model processes sets of points $\mat{y}_1^{(:)}$, $\mat{y}_2^{(:)}$, $\mat{y}_3^{(:)}$ along each spatial dimension through three independent trunk networks to compute basis function values $\tau_{n,k}^{(:)}$. The branch network generates coefficients $\beta_k$ that weight these basis functions, which are combined through outer products to efficiently represent the complete 3D temperature field as a sum of rank-1 tensors.}
    \label{fig:low_rank_sol}
\end{figure*}

Consider a three-dimensional domain discretized with $N_1$, $N_2$, and $N_3$ points along each axis. Let $\mat{y}_n^{(:)} = \{y_n^{(1)}, y_n^{(2)}, ..., y_n^{(N_n)}\}$ denote the set of coordinates along the $n$-th axis. The complete meshgrid is formed by taking the Cartesian product of these coordinate sets, resulting in $N_1 \times N_2 \times N_3$ total points. We extend Eq.~\eqref{eq:scalar_sep} to process this entire grid simultaneously:
\begin{equation}
\begin{aligned}
&G_{\boldsymbol{\theta}}(\boldsymbol{u})\left(\mat{y}_1^{(:)},\mat{y}_2^{(:)},\mat{y}_3^{(:)}\right) = \sum_{k=1}^{r} \beta_k \bigotimes_{n=1}^3 \tau_{n,k}^{(:)}\\
&=\sum_{k=1}^r b_\psi(\mathcal{E}(\boldsymbol{u}))_k \left(t^1_{\phi_1}(\mat{y}_1^{(:)})_k\otimes t^2_{\phi_2}(\mat{y}_2^{(:)})_k\otimes t^3_{\phi_3}(\mat{y}_3^{(:)})_k\right),
\end{aligned}
\label{eq:grid_sep}
\end{equation}
where $\otimes$ represents an outer product\footnote{The outer product between vectors produces a higher-dimensional tensor whose entries are all possible products between elements of the original vectors.}, and $\tau_{n,k}^{(:)} = t^n_{\phi_n}(\mat{y}_n^{(:)})_k$ is the $k$-th output of the $n$-th trunk network evaluated at all points along the $n$-th axis. As illustrated in Fig.~\ref{fig:low_rank_sol}, for each $k$, the outer product $\bigotimes_{n=1}^3 \tau_{n,k}^{(:)}$ produces a tensor of shape $N_1 \times N_2 \times N_3$ representing the contribution of the $k$-th basis function across the entire grid. This decomposition enables efficient representation of the complete 3D temperature distribution without evaluating each point individually.

\subsection{Efficient Derivative Computation}
Our separable structure, as visualized in Fig.~\ref{fig:low_rank_sol}, enables particularly efficient computation of spatial derivatives for three key reasons. First, as shown in the figure, the temperature field is decomposed into a sum of rank-1 tensors, each formed by outer products of basis functions from separate dimensions. This decomposition allows derivatives to be calculated independently along each dimension:
\begin{equation}
\begin{aligned}
\frac{\partial G_{\boldsymbol{\theta}}(\boldsymbol{u})(\mat{y}_1^{(:)},\mat{y}_2^{(:)},\mat{y}_3^{(:)})}{\partial \mat{y}_m^{(:)}} &= \sum_{k=1}^r \beta_k  \left(\bigotimes_{n\not=m} \tau_{n,k}^{(:)}\right)\otimes \frac{\partial t_{\phi_m}^m(\mat{y}_m^{(:)})_k}{\partial \mat{y}_m^{(:)}},
\end{aligned}
\label{eq:derivs_sep}
\end{equation}
where $\frac{\partial t_{\phi_m}^m(y_m^{(:)})_k}{\partial y_m^{(:)}}$ represents the derivative of the $k$-th output of the $m$-th trunk network with respect to its input coordinates. 

Second, this formulation enables the use of forward-mode automatic differentiation (AD) via Jacobian-vector products (JVPs), which is particularly efficient for our network architecture. In our separable structure, each trunk network $t_{\phi_m}^m: \mathbb{R}^{N_m} \rightarrow \mathbb{R}^{r \times N_m}$ takes $N_m$ input coordinates and produces $r$ basis functions evaluated at these coordinates. Forward-mode AD requires only $O(N_m)$ operations for JVP computation, while reverse-mode AD would require $O(r \times N_m)$ operations for vector-Jacobian products. Since $r$ (typically 32-256) is much larger than 1 in our framework, forward-mode AD provides significant computational savings.

Third, as evident from Fig.~\ref{fig:low_rank_sol}, we can reuse the outputs $\tau_{n,k}^{(:)}$ from other dimensions ($n \neq m$) in the outer product computation, avoiding redundant calculations. This feature is particularly advantageous when evaluating higher-order or mixed derivatives, as each dimension's computation remains independent. In contrast, a non-separable architecture would require computing derivatives for each point in the $N_1 \times N_2 \times N_3$ grid individually, leading to substantially higher computational complexity.

\subsection{Overall Computational Benefit}

The separable formulation offers significant advantages in physics-informed training of thermal prediction models. By allowing efficient derivative computation through forward-mode AD, our approach allows thermal analysis at higher resolutions than was previously possible with the original DeepOHeat architecture.

\section{Trustworthy Design Optimization}
In this section, we demonstrate how DeepOHeat-v1 can be integrated into a trustworthy design optimization framework for thermal optimization. \added{By ``trustworthy'', we mean predictions with quantifiable confidence and mechanisms for verification and refinement when accuracy is insufficient.}
\subsection{Thermal Optimization for Floorplan}

Thermal optimization is critical in 3D-ICs due to increased power density and complex heat dissipation paths caused by vertical stacking. These thermal challenges directly impact device reliability, performance, and lifespan, making thermal management essential for viable 3D integration technology.

The thermal optimization problem for 3D-IC floorplanning can be formulated mathematically as:
\begin{equation}
\begin{aligned}
\min_{\boldsymbol{u}} \quad & f(\mat{T}(\boldsymbol{u})) \\
 \text{subject to}
 &\quad C_i(\boldsymbol{u}) \leq 0, \quad i = 1, 2, \ldots, M,
\end{aligned}
\end{equation}
where:
\begin{itemize}
\item $\boldsymbol{u}$ represents the design configuration (component placement and properties).
\item $\mat{T}(\boldsymbol{u})$ is the temperature field for a given design $\boldsymbol{u}$.
\item $f(\mat{T}(\boldsymbol{u}))$ is the objective function, typically the peak temperature of the chip model.
\item $C_i(\boldsymbol{u}) \leq 0$ encompasses various design constraints.
\end{itemize}
As shown in the left part of Fig.~\ref{fig:trustworthy_workflow}, the optimization flow follows these general steps:
\begin{enumerate}
\item Initialize with a starting design $\boldsymbol{u}_{ini}$ and optimizer parameters.
\item For each iteration, generate new design candidates based on the current design.
\item Evaluate $\mat{T}(\boldsymbol{u})$ for each candidate design.
\item Calculate the objective function value $f(\mat{T}(\boldsymbol{u}))$.
\item Update the current design according to the optimizer.
\item Continue until termination criteria are met.
\end{enumerate}

\subsection{Confidence Estimation in Design Optimization}
This optimization problem requires solving the heat equation for each design evaluation, making it computationally intensive. While operator learning models like DeepOHeat-v1 can significantly accelerate this process, they introduce new challenges for design reliability. Due to the black-box nature of neural networks, there is no guarantee that the predicted result of the operator learning model is accurate enough for every design case. In fact, DeepOHeat-v1 can have poor accuracy for some corner design cases. Consequently, an optimization solver may provide a low-quality solution once we use DeepOHeat-v1 directly in every optimization step.

In order to address this trustworthiness issue, we propose to assess the confidence of a DeepOHeat-v1 prediction at low cost and {\it without explicitly solving the underlying PDEs}. Specifically, assume that the PDE is first discretized (e.g. via a finite difference method) into a linear system:
\begin{equation}
\mat{A}\mat{T} = \mat{b}.
\label{eq:linear_system}
\end{equation}
Here the system matrix $\mat{A}$ and load vector $\mat{b}$ are functions of the design configuration $\boldsymbol{u}$, which may include thermal conductivity variations, power distribution patterns, and boundary conditions. We do not intend to solve for $\mat{b}$ due to the high computational cost. Instead, we can easily obtain $\mat{T}_0$, which is the predicted temperature at all discretization points using DeepOHeat-v1. The residual of the linear system provides a natural and efficient measure of solution quality:
\begin{equation}
\label{eq:confidence}
r = \frac{\|\mat{A}\mat{T}_0 - \mat{b}\|}{\|\mat{b}\|}.
\end{equation}
This metric directly measures how well the predicted temperature field satisfies the discretized thermal equations without requiring the true solution. A small residual ($r < \alpha$) indicates that the prediction closely satisfies the governing equations, while a large residual suggests potential inaccuracies.

\subsection{A Trustworthy Hybrid Optimization Framework}
\begin{figure*}
    \centering
    \includegraphics[width=0.7\linewidth]{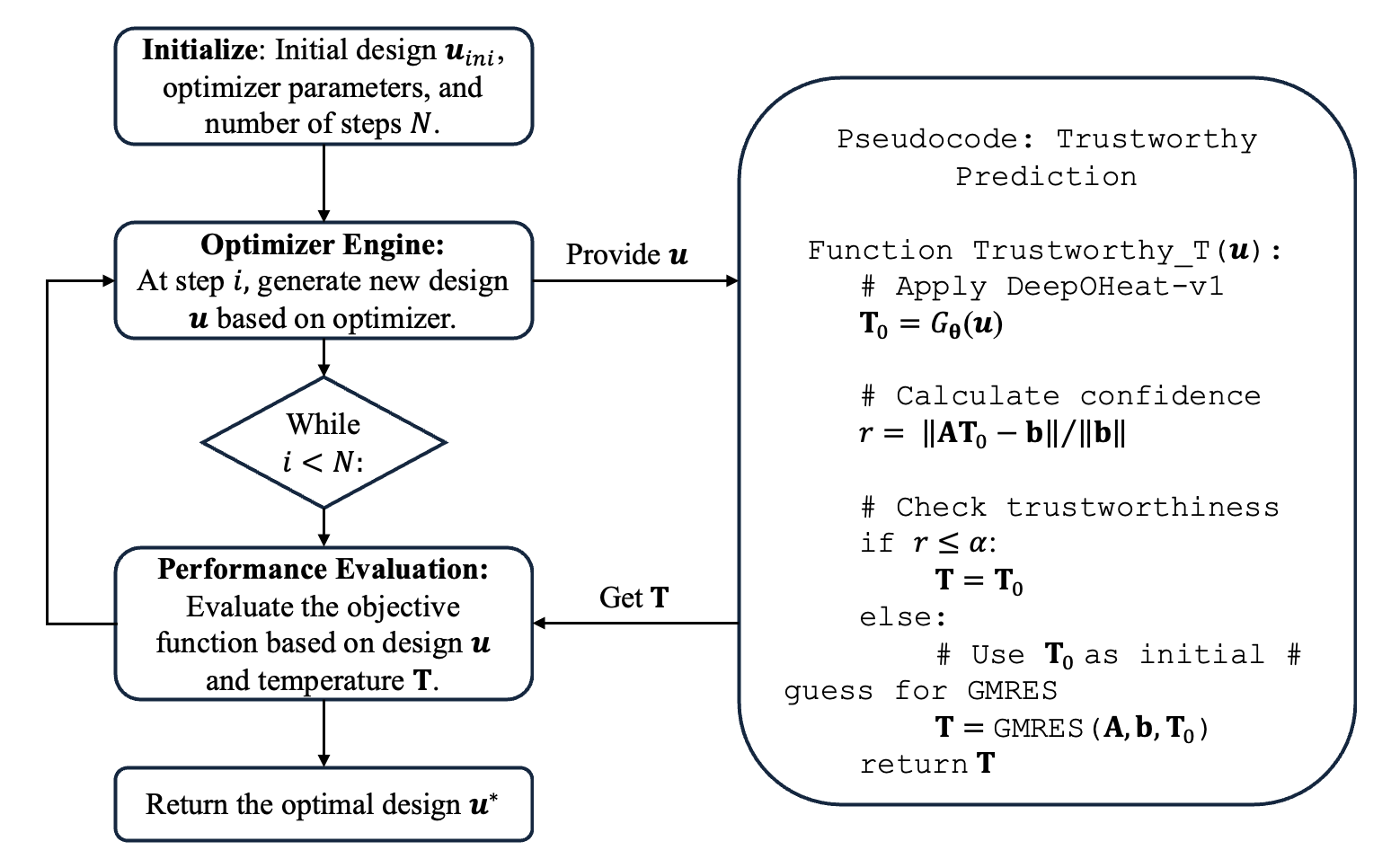}
    \caption{Hybrid thermal optimization framework for 3D-IC design. The framework consists of two nested loops: (1) an outer optimizer loop (left) that explores the design space using any optimization algorithm of choice; and (2) an inner trustworthy prediction loop (right) that ensures reliable thermal evaluation by applying DeepOHeat-v1 and adaptively refining predictions with GMRES when the residual exceeds a predefined threshold. }
    \label{fig:trustworthy_workflow}
\end{figure*}

Based on the confidence estimator above, we now develop a hybrid optimization framework that combines the efficiency of operator learning with the reliability of numerical methods, as illustrated in Fig. \ref{fig:trustworthy_workflow}. The right side of the figure shows our inner loop, which implements an adaptive solution strategy for each design evaluation:
\begin{enumerate}
\item For each new design configuration $\boldsymbol{u}$, obtain an initial prediction through the operator learning model:
\begin{equation}
\mat{T}_0 = G_{\boldsymbol{\theta}}(\boldsymbol{u}).
\label{eq:initial_guess}
\end{equation}
\item Evaluate prediction trustworthiness through the relative residual described in Eq.~\eqref{eq:confidence}.
\item If $r < \alpha$, we accept $\mat{T}_0$ directly; otherwise, we refine the solution using GMRES with $\mat{T}_0$ as an {\it initial guess}. Although $\mat{T}_0$ is not accurate enough, it can play as an extremely high-quality initial guess of GMRES. Our implementation shows that this initial guess can reduce the number of GMRES iteration from thousands to a few dozens. 
\end{enumerate}

Only a small portion of the optimization steps need GMRES-based solution refinement. Furthermore, only a small number of GMRES iterations are required to solve each linear system. As a result, the entire optimization framework has a low computational cost, which is only marginally more expensive than using operator learning alone while providing the reliability of high-fidelity numerical methods.

In this work, we use simulated annealing \cite{van1987simulated} as the outer-loop optimizer (see Algorithm \ref{alg:simulated_annealing} in Appendix \ref{app:simulated_annealing}). However, our adaptive solution method can be integrated with any other outer-loop optimizer, including but not limited to gradient descent, reinforcement learning, and Bayesian optimization. 

\begin{figure}
    \centering
    \includegraphics[width=\linewidth]{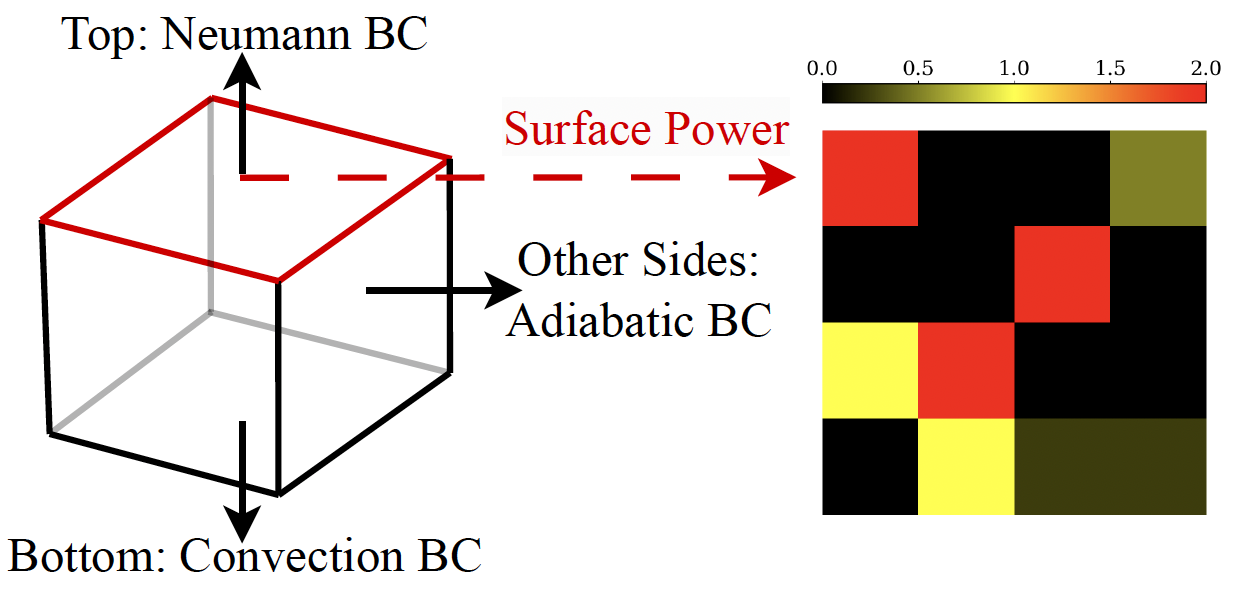}
    \caption{Chip geometry used in thermal simulations. Left: Conceptual single-cuboid model with isotropic material properties. Right: power map applied to the top surface.}
    \label{fig:surface geometry}
\end{figure}

\section{Experiments}
We evaluate our enhanced framework through comprehensive experiments, systematically demonstrating how each proposed improvement addresses the limitations of DeepOHeat. \added{Our experiments use $1\text{mm} \times 1\text{mm}$ chip dimensions in the $xy$-direction, following the original DeepOHeat paper for fair comparison, while our KAN and separable training improvements are architecture-independent and provide greater benefits at larger scales.} First, we assess the framework on a 2D power map thermal simulation benchmark to validate both prediction accuracy and computational efficiency improvements. 
Second, we validate our trustworthy optimization workflow in a floorplan design scenario, where reliable temperature prediction is crucial for finding optimal component placements. Throughout our experiments, we compare following models:
\begin{itemize}
    \item \emph{DeepOHeat}: The baseline operator learning model from \cite{liu2023deepoheat} using MLP-based trunk networks.
    \item \emph{DeepOHeat + KAN}: An enhanced variant that replaces the MLP in the trunk network with Kolmogorov-Arnold Networks (KAN) for improved representational capacity.
    \item \emph{DeepOHeat-v1}: Our fully enhanced model implementing both KAN-based trunk nets and separable training.
    \item \emph{DeepOHeat + ST}: An ablation variant incorporating only our separable training approach with the original MLP architecture (introduced in Section \ref{subsec:3D-arch}).
\end{itemize}

\begin{figure*}[t]
    \centering
    \includegraphics[width=0.7\textwidth]{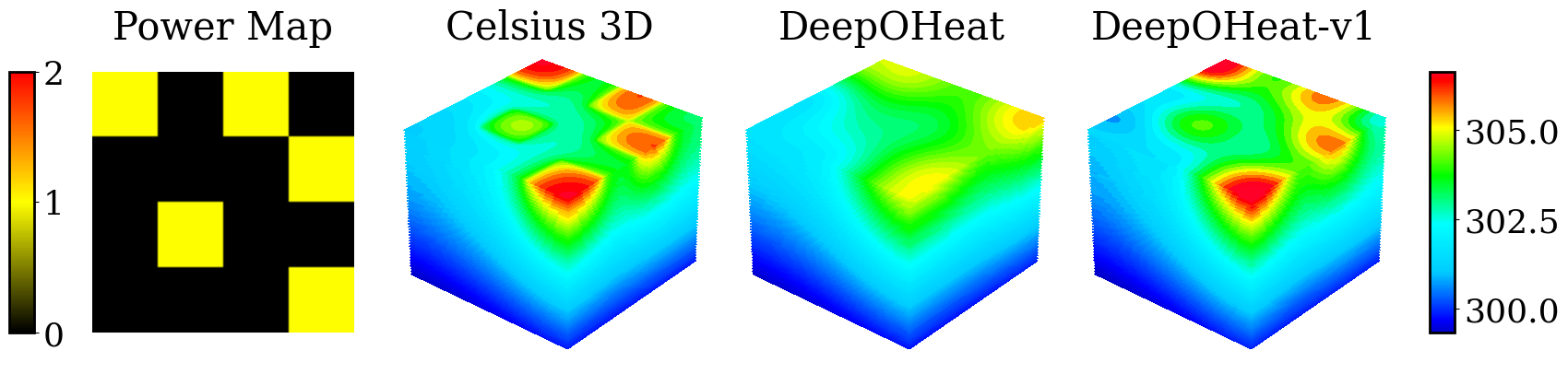}
    \caption{Comparison of temperature field predictions for a representative multi-hotspot power map. From left to right: input power map showing the complex thermal pattern, reference solution by Celsius 3D, temperature predictions by DeepOHeat, and temperature predictions by DeepOHeat-v1. The visualization demonstrates DeepOHeat-v1's superior ability to capture multi-scale thermal features, particularly at hotspot regions where the baseline DeepOHeat model fails to capture more detailed patterns.}
    \label{fig:2dpower_pred}
\end{figure*}

\begin{figure*}[t]
    \centering
    \includegraphics[width=0.6\textwidth]{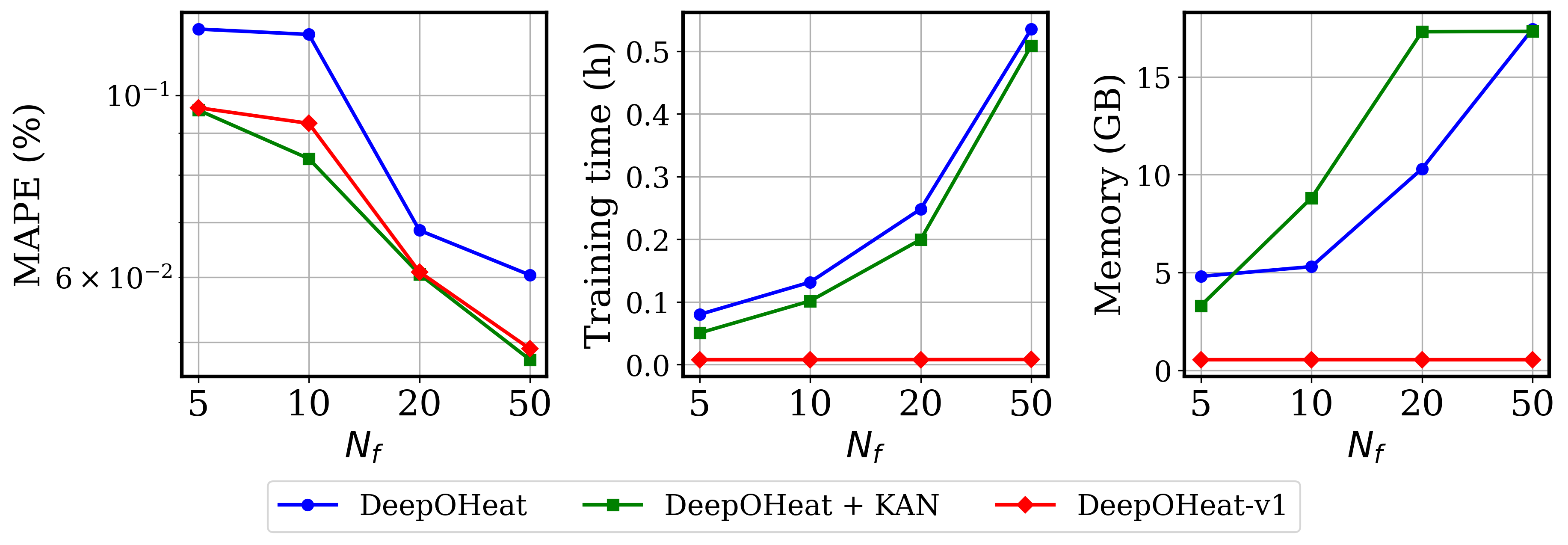}
    \caption{Performance comparison of three models in terms of MAPE, GPU memory usage, and training time for different $N_f$.}
    \label{fig:2dpower_varynf}
\end{figure*}

\subsection{2D Power Map on the Top Surface} \label{example1}
\subsubsection{Problem Formulation and Implementation}
Following DeepOHeat \cite{liu2023deepoheat}, we consider a single-cuboid chip model with dimensions $1\text{mm} \times 1\text{mm} \times 0.5\text{mm}$, discretized on a $21\times 21 \times 11$ mesh grid (Fig.~\ref{fig:surface geometry}). The power map is defined on a $20 \times 20$ tiling of the top surface, with each tile's power level scaled by $0.00625(mW)$ and interpolated onto the $21 \times 21$ computational grid. We maintain adiabatic conditions on the side surfaces and convection conditions on the bottom surface ($\text{HTC}=500 W/(m^2 K)$, $T_{\rm amb}=298.15 (K)$), with homogeneous thermal conductivity $k = 0.1 W/(mK)$.

All models share the same branch network architecture (9 layers, 256 neurons per layer) processing the 441-dimensional flattened power map input. For trunk networks, the baseline DeepOHeat employs a single 6-layer MLP (128 nodes/layer) with Fourier feature mapping ($\mu=0$, $\sigma=2\pi$) and swish activation. DeepOHeat + KAN maintains the single trunk network architecture but replaces the MLP with a 4-layer order-3 Chebyshev KAN (64 nodes/layer) to keep the parameter budget comparable. DeepOHeat-v1 implements our separable training approach with three independent trunk networks (one per spatial dimension), each having the same KAN architecture as DeepOHeat + KAN. The physics-informed loss is evaluated on the full computational mesh. To generate diverse training data, we sample 50 power maps per iteration from a two-dimensional standard Gaussian random field (GRF) with length scale parameter of 0.3. All implementations are built with JAX for efficient computation. Models are trained for 10,000 iterations using ADAM optimizer (learning rate 1e-3, decayed by 0.9$\times$ every 500 iterations) on an NVIDIA RTX 3090 GPU.

\subsubsection{Results Analysis}
Table \ref{tab:2d_comparison} presents performance comparisons with $N_f=50$ sampled functions per iteration. The reported mean absolute percentage error (MAPE) values are calculated and averaged based on ten representative power maps from the original DeepOHeat paper. \added{Appendix \ref{appendix:2d_all_cases} provides visualizations of all 10 test cases.} \added{Importantly, these evaluation power maps consist of structured heat blocks that differ significantly from the Gaussian Random Field (GRF) patterns used during training.} DeepOHeat + KAN demonstrates superior prediction accuracy, reducing MAPE from 0.060\% to 0.048\% compared to the baseline. This $1.25\times$ reduction in error validates that KAN's learnable activation functions better capture multi-scale thermal patterns. The computational cost remains comparable to the baseline (0.51h vs 0.53h, 17.32GB vs 17.44GB), indicating the improved accuracy comes without additional overhead.

DeepOHeat-v1 maintains the enhanced prediction capability (0.049\% MAPE) while drastically reducing computational requirements through separable training. Training time decreases by $62\times$ (from 0.51h to 30.24s) and memory usage by $31\times$ (from 17.32GB to 0.56GB), demonstrating effective mitigation of the computational bottleneck.

Fig.~\ref{fig:2dpower_pred} illustrates the practical impact of these improvements with one representative multi-hotspot power distribution from our test set. Visual comparison between the Celsius 3D reference solution and model predictions reveals that DeepOHeat struggles with accurately capturing sharp temperature gradients around hotspots, particularly in the central and corner regions. In contrast, DeepOHeat-v1 produces temperature fields that more closely match the reference solution, with better-defined hotspot shapes and intensities. This demonstrates KAN's superior ability to represent the multi-scale features inherent in thermal fields, particularly when combined with our separable training approach.

Fig. \ref{fig:2dpower_varynf} further analyzes model behavior across varying $N_f$ values, revealing several key insights:
\begin{itemize}
    \item \emph{Architectural Impact}: DeepOHeat + KAN consistently achieves lower MAPE than the baseline across all $N_f$ values, with the improvement becoming more pronounced at larger batch sizes. This demonstrates KAN's robust advantage in capturing thermal patterns.
    \item \emph{Training Efficiency}: Although both baseline and KAN-only variants show rapid growth in computational requirements with increasing $N_f$, DeepOHeat-v1 maintains nearly constant memory usage (0.56GB) and sub-minute training time. This scalability is crucial for practical applications that require larger training batches.
    \item \emph{Synergistic Benefits}: DeepOHeat-v1 combines the accuracy of KAN with the efficiency of separable training, achieving optimal performance in all metrics. 
\end{itemize}

\begin{table}[t]
\caption{Performance Comparison for 2D Power Map on Top Surface. }
\label{tab:2d_comparison}
\centering
\resizebox{\columnwidth}{!}{
\begin{tabular}{lccc}
\hline
\textbf{Model} & \textbf{MAPE (\%)}  & \textbf{Training Time (h)} & \textbf{GPU Memory (GB)} \\
\hline
DeepOHeat & 0.060 & 0.53 & 17.44  \\
DeepOHeat + KAN & 0.048 & 0.51 & 17.32  \\
DeepOHeat-v1 & 0.049 & 0.0084 & 0.56  \\
\hline
\end{tabular}
}
\end{table}

\subsection{Volumetric Power Distribution in Three-Layer Architecture}
\label{subsec:3D-arch}

In this section, we evaluate our framework in a more challenging thermal simulation scenario involving volumetric power distributions in a complex 3D chip architecture. This experiment presents a significant computational challenge: high-resolution 3D thermal analysis with volumetric power distribution across multiple materials.

\subsubsection{Problem Setup}
\begin{figure}
    \centering
    \includegraphics[width=\linewidth]{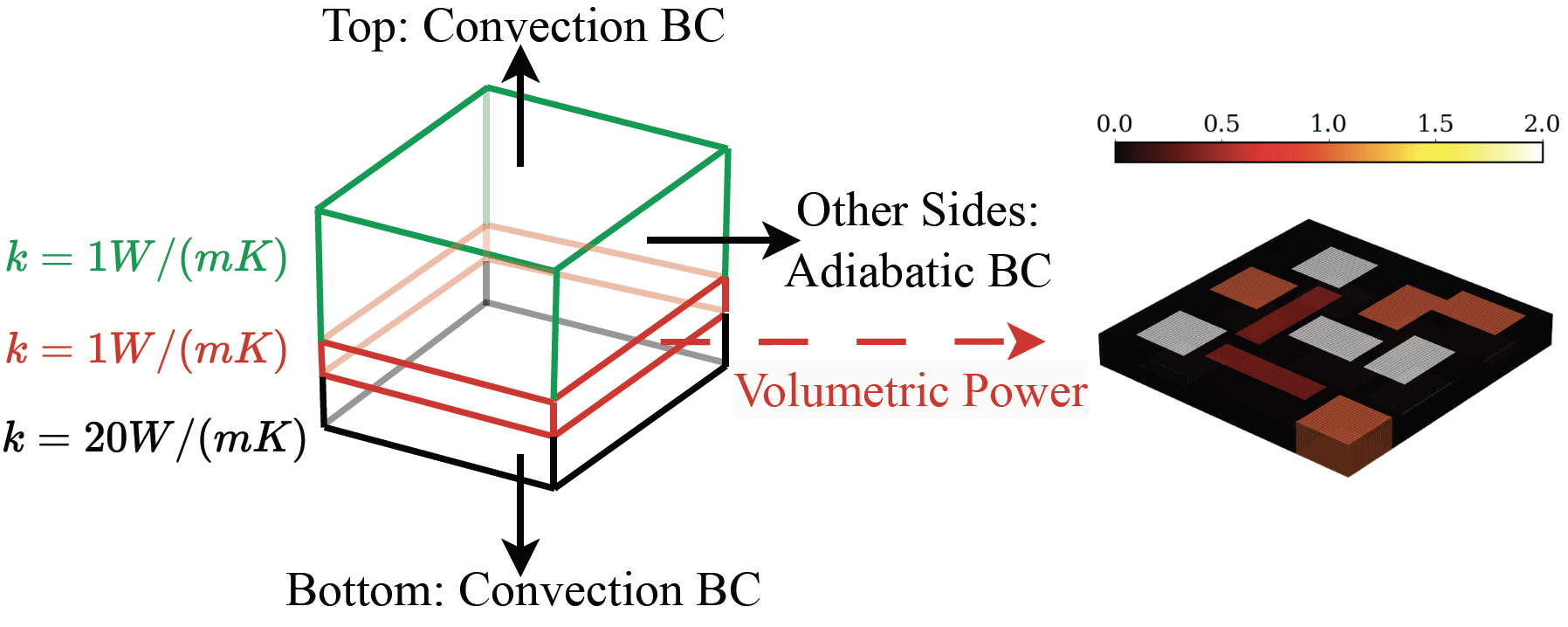}
    \caption{3-D chip geometry: (left) schematic showing top (green), middle (red), and bottom (gray) layers; (right) a volumetric power map (heat sources) applied to the middle layer.}
    \label{fig:multilayer_chip}
\end{figure}

\begin{figure*}[t]
    \centering
\includegraphics[width=0.7\textwidth]{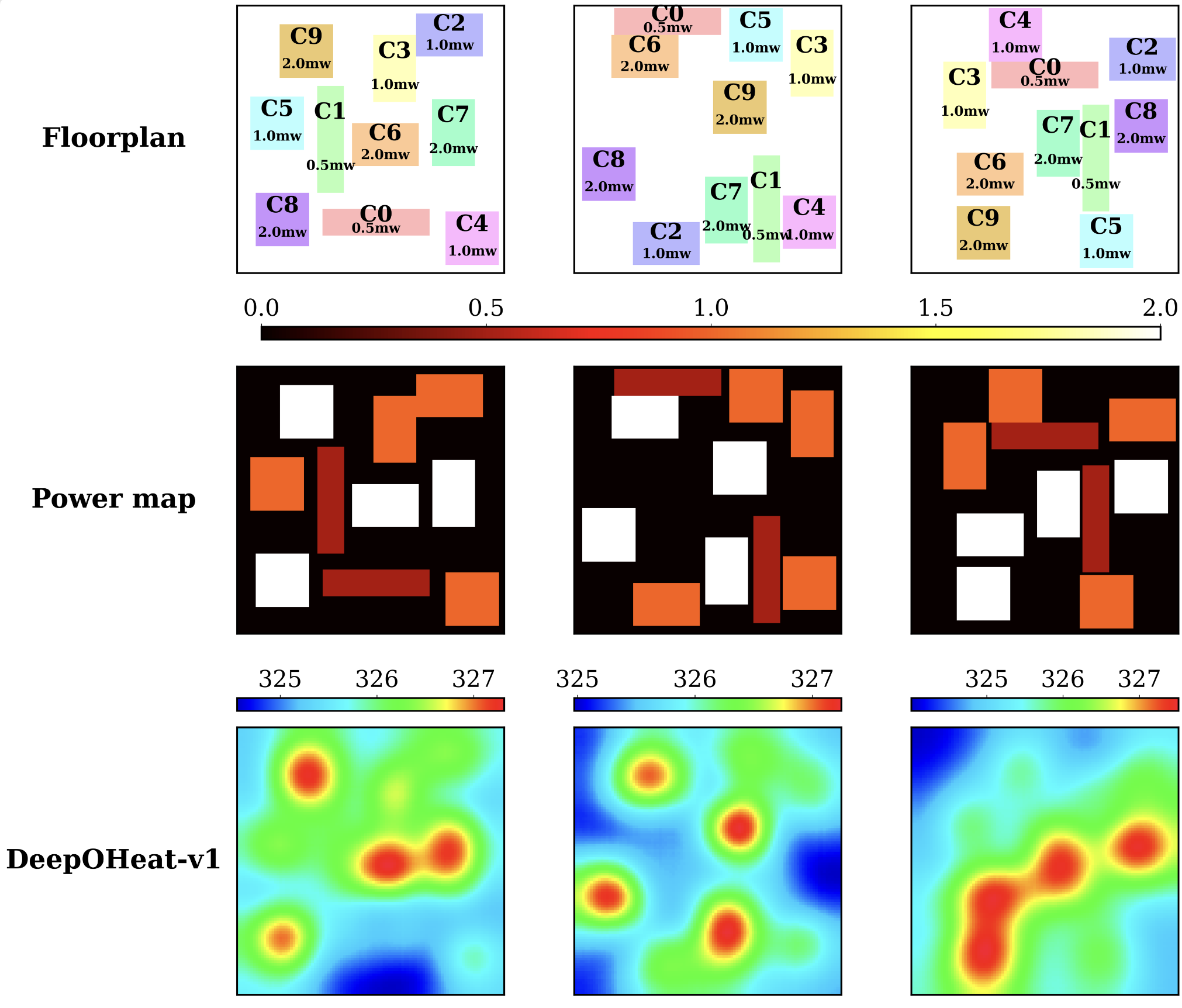}
    \caption{Three example floorplan configurations (a-c) and their thermal responses. Top row: spatial layout of ten components (C0-C9) with different power densities; Middle row: resulting power maps on the computational grid; Bottom row: temperature distributions predicted by DeepOHeat-v1 at the 13th vertical slice (z = 0.12mm).}
    \label{fig:example_floorplans}
\end{figure*}

 We consider a three-layer chip model ($1\text{mm} \times 1\text{mm} \times 0.55\text{mm}$) discretized on a $101 \times 101 \times 56$ computational mesh (Fig.~\ref{fig:multilayer_chip}). The model has thermal conductivities of $1 W/(mK)$ (top layer, 0.4mm), $1 W/(mK)$ (middle layer, 0.05mm), and $20 W/(mK)$ (bottom layer, 0.1mm), with adiabatic side boundaries and convection on the top/bottom ($\text{HTC}=500 W/(m^2 K)$, $T_{\rm amb}=298.15 (K)$). The power distribution is defined by ten rectangular components (C0-C9) placed in the middle layer. Each component occupies a fixed area (400 tiles on a $100 \times 100$ grid) but with varying power: $0.5(mW)$ (C0, C1), $1.0(mW)$ (C2-C5), and $2.0(mW)$ (C6-C9). As shown in Fig. \ref{fig:example_floorplans}, different spatial arrangements of these components result in distinct volumetric power patterns, creating diverse thermal loading scenarios.

\subsubsection{Surrogate Model Training and Evaluation}
The high-resolution mesh ($101 \times 101 \times 56$)\added{, representing a spatial resolution of approximately 10 \textmu m per grid point,} presents a significant computational challenge, requiring more than 570K spatial points during the physics-informed training. \added{During training, we minimize the physics loss across this entire fine mesh, enforcing thermal governing equations and boundary conditions at high spatial resolution. This high-resolution physics-informed training produces more accurate surrogate models compared to training on coarser grids, and was computationally infeasible with the original DeepOHeat due to memory limitations.} Only models incorporating separable training—DeepOHeat + ST and DeepOHeat-v1—can be effectively trained on an NVIDIA RTX 3090 GPU.

We maintain the same network architectures as in Section \ref{example1}, with the only difference being the branch network input dimension increased to 10,201 to handle the $101 \times 101$ power map in the middle layer. During training, we generate random floorplan configurations by stochastically placing the ten components, allowing overlap for efficient sampling. Each model is trained for 100,000 iterations using ADAM optimizer (learning rate 1e-3, decayed by 0.9$\times$ every 1,000 iterations), with 50 floorplans sampled per iteration. The physics-informed loss enforces thermal governing equations and boundary conditions across the entire computational domain.

For performance evaluation, we generate 100 random test floorplans following the same sampling procedure as the training data. The prediction accuracy is measured against reference solutions computed using Generalized Minimal Residual Method (GMRES) with algebraic multigrid preconditioning. Table \ref{tab:volumetric_simulation} presents a detailed performance comparison. DeepOHeat-v1 achieves dramatically improved accuracy, reducing MAPE from 0.22\% to 0.035\%—a 6.29$\times$ reduction in error—while maintaining similar computational efficiency (approximately 0.5 hours of training time and 12GB memory usage). This significant improvement highlights KAN's superior capability in capturing complex 3D thermal patterns, while separable training enables scalability to high resolutions.

Fig.~\ref{fig:example_floorplans} shows example temperature predictions for three different component arrangements. The temperature distributions at the 13th vertical slice (z = 0.12mm) show the formation of hotspots at locations with high-power components (e.g., C6-C9), and thermal coupling between neighboring components.

\begin{table}[t]
\caption{Performance Comparison of DeepOHeat-v1 Variants for Volumetric Power in Three-Layer Chip Architecture}
\label{tab:volumetric_simulation}
\centering
\resizebox{\columnwidth}{!}{
\begin{tabular}{lccc}
\hline
\textbf{Model} & \textbf{MAPE (\%)} &  \textbf{Training Time (h)} & \textbf{GPU Memory (GB)} \\
\hline
DeepOHeat + ST & 0.22 & 0.50 & 12.09  \\
DeepOHeat-v1 & 0.035 & 0.52 & 12.10  \\
\hline
\end{tabular}
}
\end{table}

\subsection{Thermal Optimization for Floorplan}
Finally, we use our DeepOHeat-v1 to optimize the floorplan of the three-layer design described in Section~\ref{subsec:3D-arch}. We first evaluate our confidence estimator and solution refinement method. Then we evaluate the performance of the entire floorplan optimizer.

\subsubsection{Confidence estimator and solution refinement}
We evaluate three solution methods for thermal analysis using the left configuration of Fig. \ref{fig:example_floorplans}:
\begin{itemize}
    \item GMRES attempts to solve the thermal linear system directly from a random initial guess, demonstrating the challenge of the thermal problem and the importance of a good initial guess for convergence. 
    \item The operator learning model (DeepOHeat-v1) provides direct temperature predictions through the trained operator model without solving the linear system.
   \item Our hybrid approach first obtains an initial prediction from DeepOHeat-v1, and then refines it using GMRES when the prediction residual exceeds the threshold. This shows how a good initial guess from the operator model significantly accelerates GMRES convergence.
\end{itemize}

\begin{figure}[t]
    \centering
    \includegraphics[width=0.9\linewidth]{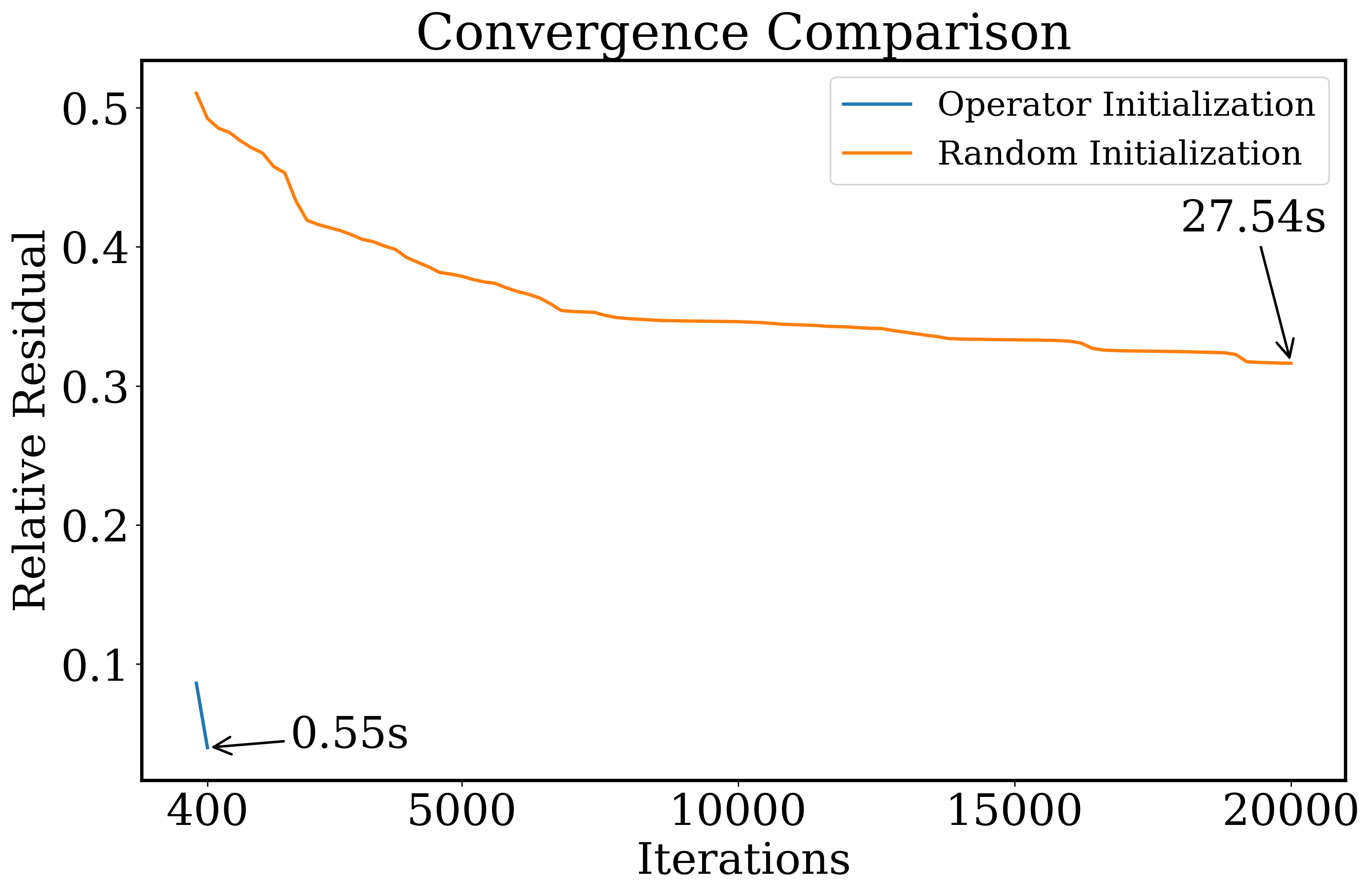}
    \caption{Convergence comparison between GMRES with different initialization strategies. The operator initialization (blue) leverages DeepOHeat-v1 predictions as the initial guess, enabling rapid convergence in 0.55s. Random initialization (orange) requires 27.54s and still fails to reach acceptable accuracy after 20,000 iterations.}
    \label{fig:convergence_comparison}
\end{figure}

All GMRES implementations utilize CuPy \cite{nishino2017cupy} for GPU acceleration, with parameters set to: restart = 200, maximum iterations = 20,000, and tolerance = 0.05. Fig. \ref{fig:convergence_comparison} compares the convergence behavior of GMRES with different initializations. Standard GMRES with a random initial guess converges slowly, taking 27.54s to reach a modest residual level even after 20,000 iterations. In contrast, when initialized with our operator learning prediction, GMRES converges rapidly to a relative residual of 0.04 in just 0.55s. This 50× speedup demonstrates the effectiveness of neural network predictions as initial guesses for iterative solvers.

\begin{figure*}[t]
    \centering
    \includegraphics[width=0.7\textwidth]{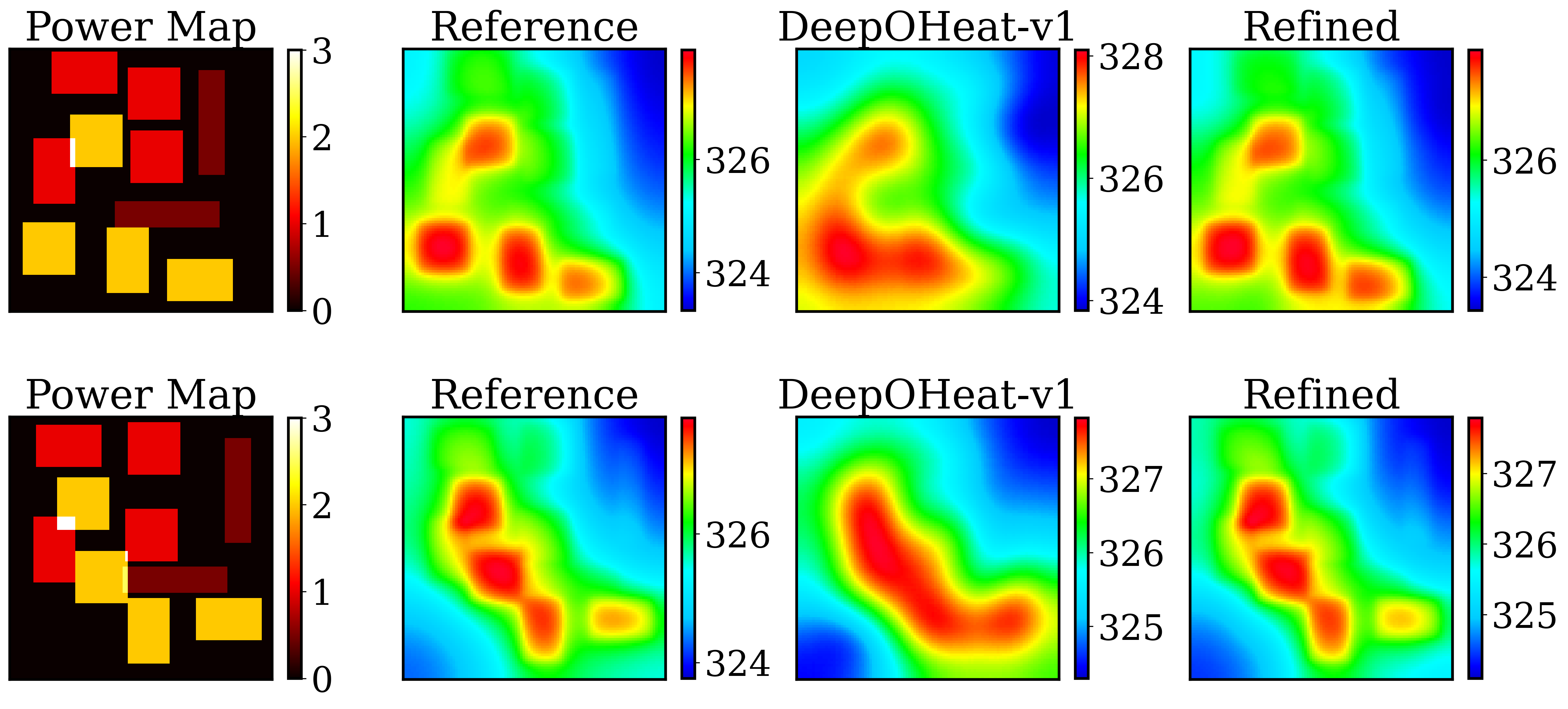}
    \caption{Thermal prediction reliability assessment during optimization. Top and bottom rows show two cases where operator model predictions required refinement. For each case: power distribution (left), reference solution (second), DeepOHeat-v1 prediction (third), and hybrid method refinement (right). The refined solutions correct the inaccuracies in the neural network predictions.}
    \label{fig:refined_pred}
\end{figure*}

Fig. \ref{fig:refined_pred} shows two representative cases where the predictions from operator learning require refinement during optimization. While the operator model captures general thermal patterns, the hybrid refinement successfully corrects local prediction errors to match the reference solution. This is particularly evident in areas with closely positioned high-power components, where thermal coupling effects create complex temperature gradients that are challenging to predict perfectly.

\subsubsection{Overall Performance of Thermal Optimization for Floorplan}
We evaluate our hybrid framework in a thermal optimization task in floorplanning. The optimization objective is to minimize the peak temperature of the chip model by determining optimal component placements while avoiding overlaps through adding a soft penalty term to the objective function.We implement simulated annealing using Optuna \cite{akiba2019optuna} with 1,000 optimization steps, an initial SA temperature of 0.5, a cooling rate of 0.997, and a base neighborhood size of 25 (as detailed in Algorithm \ref{alg:simulated_annealing} in Appendix \ref{app:simulated_annealing}). For the hybrid framework's residual threshold, we set $\alpha$ to 9.

\begin{figure}
    \centering
    \includegraphics[width=0.9\linewidth]{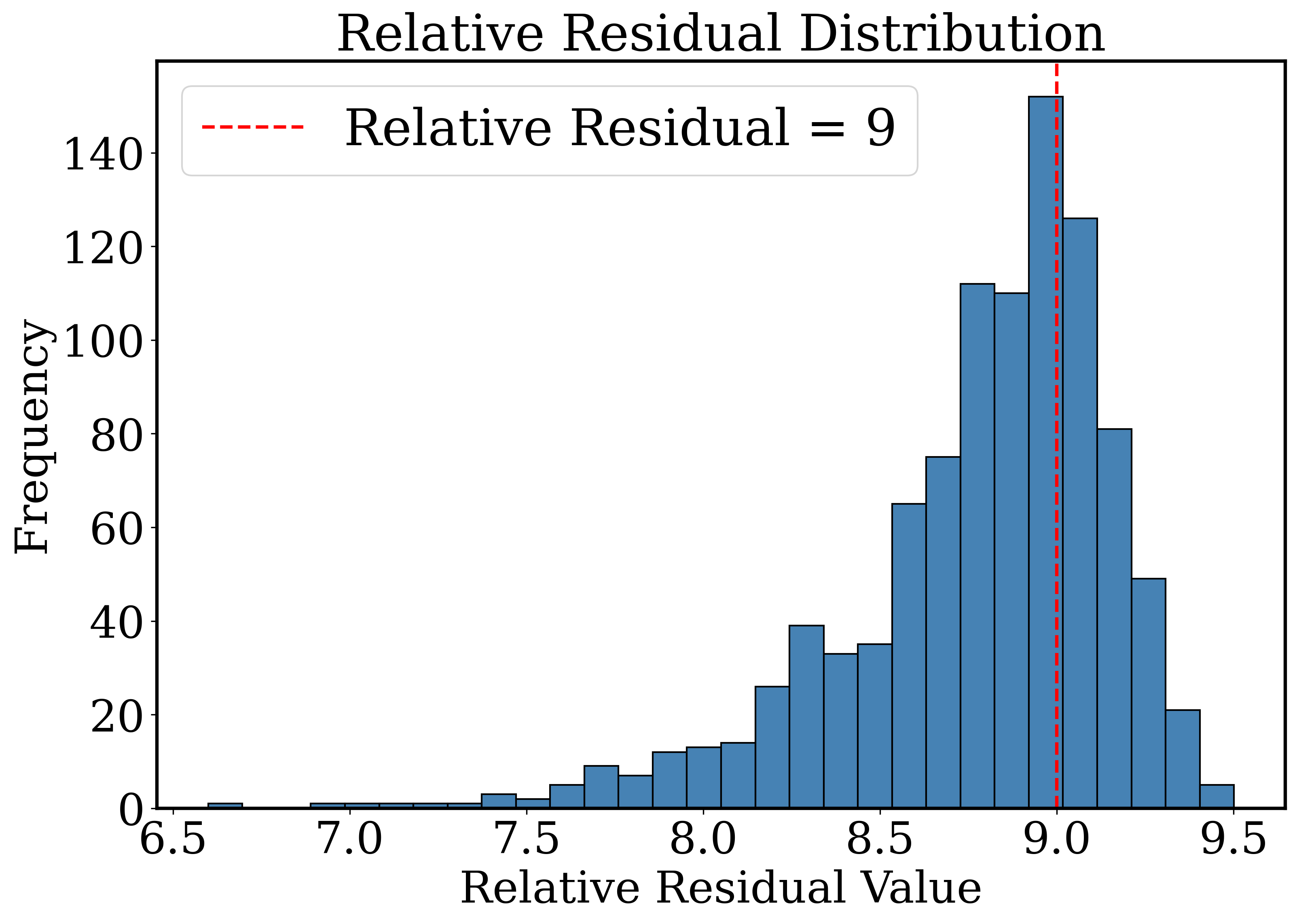}
    \caption{Distribution of relative residual values from DeepOHeat-v1 predictions across 1,000 optimization trials.}
    \label{fig:error dist}
\end{figure}

Fig. \ref{fig:error dist} shows the distribution of relative residual values across all design evaluations in our optimization process. With our selected threshold of $\alpha = 9$ (indicated by the vertical dashed line), we observe that approximately 30\% of the design evaluations triggered the GMRES refinement step. This analysis confirms that while most predictions from DeepOHeat-v1 were sufficiently accurate, a significant portion required additional refinement to ensure solution reliability.

Table \ref{tab:floorplan_optimization} compares the three thermal evaluation approaches. The results demonstrate the effectiveness of our hybrid framework in balancing accuracy and efficiency. GMRES with algebraic multigrid preconditioning serves as our reference, providing the most accurate solution but at the highest computational cost. Pure operator learning achieves significant speedup (0.05h vs. 4.94h), but does not offer a guarantee of trustworthiness during the optimization process. Our hybrid approach addresses this concern by monitoring the quality of the prediction through residual verification, triggering non-preconditioned GMRES refinement in 299 out of 1,000 optimization steps. By ensuring reliable temperature evaluation throughout the optimization, the hybrid approach not only provides solution guarantees but also achieves better design quality (327.31$(K)$ vs. 328.50$(K)$ maximum temperature) compared to pure operator learning, while closely matching the reference solution (327.18$(K)$). This improvement comes with minimal computational overhead (0.02h additional optimization time), demonstrating the value of selective refinement for trustworthy design optimization.

\begin{figure*}[t]
    \centering
    \includegraphics[width=0.7\textwidth]{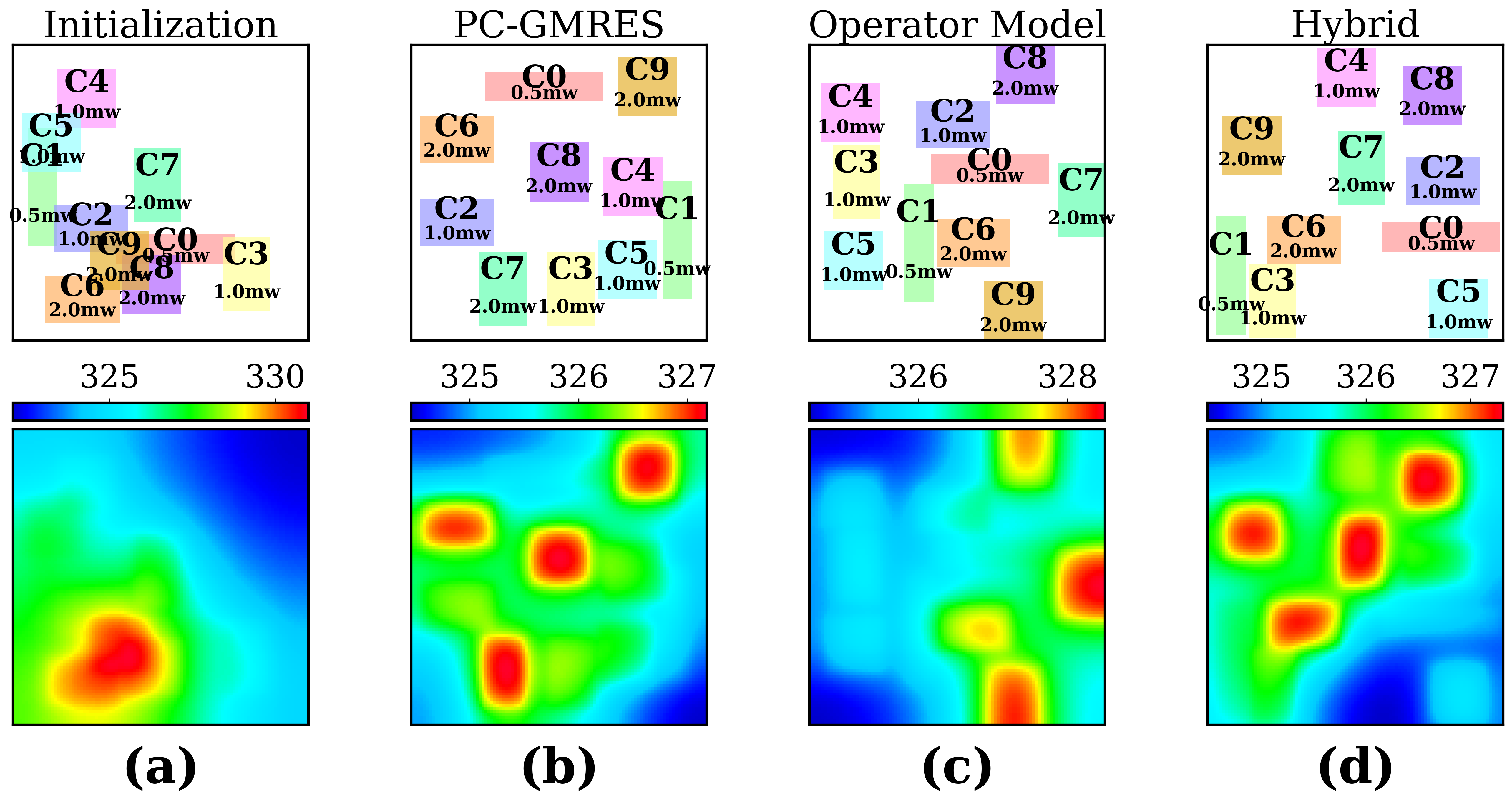}
    \caption{Floorplan configurations (top) and their corresponding temperature distributions (bottom). (a) Initial design before optimization, (b) Optimized design using PC-GMRES with algebraic multigrid preconditioning, (c) Optimized design using operator model only, and (d) Optimized design using hybrid approach.}
    \label{fig:optimized_floorplans}
\end{figure*}

The final optimized floorplans in Fig. \ref{fig:optimized_floorplans} show that our hybrid approach produces design solutions of similar quality to non-machine learning method while reducing computating timet by $70\times$.

\begin{table}[t]
\caption{Floorplan Optimization Performance Analysis Using Different Thermal Evaluation Methods}
\label{tab:floorplan_optimization}
\centering
\begin{tabular}{lcc}
\hline
\textbf{Evaluator in Optimization} & \textbf{Max Temp ($K$)} & \textbf{Opt. Time (h)} \\
\hline
GMRES w/ preconditioning & 327.18 & 4.94 \\
Operator learning only & 328.50 & 0.05 \\
Proposed hybrid solver & 327.31 & 0.07 \\
\hline
\end{tabular}
\end{table}

\section{Conclusion}
This work has presented DeepOHeat-v1 to significantly enhance the original DeepOHeat framework through three key innovations. By replacing MLP trunk networks with Kolmogorov-Arnold Networks, we have enabled adaptive learning of multi-scale thermal features, leading to substantial improvements in prediction accuracy for both surface and volumetric power distributions. The employment of separable training has overcome the computational bottleneck in physics-informed training, making high-resolution thermal analysis feasible on standard GPU hardware. Most importantly, our confidence estimator and hybrid optimization framework transform operator learning from a fast but potentially unreliable predictor into a trustworthy design optimization tool for thermal optimization in 3D-IC design. The selective refinement strategy effectively balances efficiency and reliability, achieving accuracy comparable to an optimization engine using \replaced{high-fidelity}{gold}reference mesh-based PDE solvers while obtaining significant speedup. This capability is particularly valuable for practical 3D-IC design, where reliable thermal evaluation is crucial to finding optimal designs in design space exploration and signal/power integrity sign-off.

These advances open new possibilities for efficient thermal optimization in increasingly complex 3D-IC and chiplet design. \added{Several important extensions can build upon this foundation. First, extending to transient thermal analysis by incorporating time as an additional input dimension to the trunk networks. Second, addressing full 3D-IC specific challenges including through-silicon via (TSV) thermal modeling, inter-die thermal coupling, and thermal-aware placement considering packaging constraints. Third, validation on larger chip sizes representative of modern designs. Fourth, and critically, our current test cases demonstrate modest temperature gradients; validation on modern high-performance designs with extreme gradients is essential to fully assess the method's capabilities. Finally, extending our framework to address more practical thermal-aware floorplan optimization problems, including considerations for power delivery networks, timing constraints, and multi-objective optimization scenarios that balance thermal performance with other critical design metrics.}
\begin{figure*}[t]
\centering
\includegraphics[width=\linewidth]{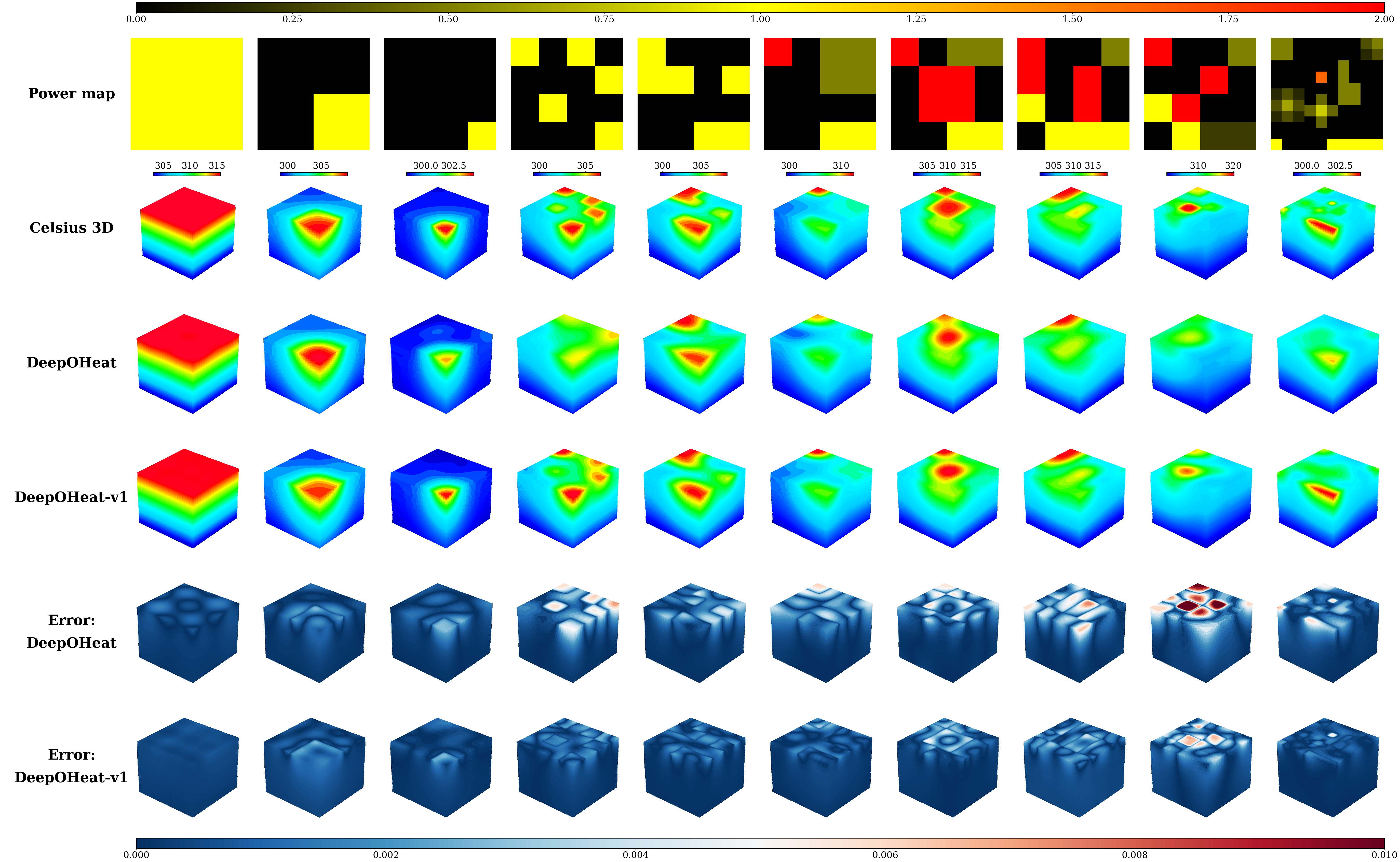}
\caption{All 10 test power maps from the original DeepOHeat paper used in Section VI.A evaluation. Top row: power distributions. Rows 2-4: temperature predictions from Celsius 3D (reference), DeepOHeat, and DeepOHeat-v1. Bottom rows: error maps for DeepOHeat and DeepOHeat-v1.}
\label{fig:appendix_2d_all_cases}
\end{figure*}

\bibliographystyle{IEEEtran}
\bibliography{main}

% Generated by IEEEtran.bst, version: 1.14 (2015/08/26)
\begin{thebibliography}{10}
\providecommand{\url}[1]{#1}
\csname url@samestyle\endcsname
\providecommand{\newblock}{\relax}
\providecommand{\bibinfo}[2]{#2}
\providecommand{\BIBentrySTDinterwordspacing}{\spaceskip=0pt\relax}
\providecommand{\BIBentryALTinterwordstretchfactor}{4}
\providecommand{\BIBentryALTinterwordspacing}{\spaceskip=\fontdimen2\font plus
\BIBentryALTinterwordstretchfactor\fontdimen3\font minus \fontdimen4\font\relax}
\providecommand{\BIBforeignlanguage}[2]{{%
\expandafter\ifx\csname l@#1\endcsname\relax
\typeout{** WARNING: IEEEtran.bst: No hyphenation pattern has been}%
\typeout{** loaded for the language `#1'. Using the pattern for}%
\typeout{** the default language instead.}%
\else
\language=\csname l@#1\endcsname
\fi
#2}}
\providecommand{\BIBdecl}{\relax}
\BIBdecl

\bibitem{liu2023deepoheat}
Z.~Liu, Y.~Li, J.~Hu, X.~Yu, S.~Shiau, X.~Ai, Z.~Zeng, and Z.~Zhang, ``{DeepOHeat:} operator learning-based ultra-fast thermal simulation in 3d-ic design,'' in \emph{2023 60th ACM/IEEE Design Automation Conference (DAC)}.\hskip 1em plus 0.5em minus 0.4em\relax IEEE, 2023, pp. 1--6.

\bibitem{3dic_benefits}
J.~H. Lau, ``{TSV} manufacturing yield and hidden costs for {3D} {IC} integration,'' in \emph{2010 Proceedings 60th electronic components and technology conference (ECTC)}.\hskip 1em plus 0.5em minus 0.4em\relax IEEE, 2010, pp. 1031--1042.

\bibitem{3dic_review}
B.~Wu and A.~Kumar, ``Extreme ultraviolet lithography and three dimensional integrated circuit—a review,'' \emph{Applied Physics Reviews}, vol.~1, no.~1, 2014.

\bibitem{optimal}
H.~Delaram, A.~Dastfan, and M.~Norouzi, ``Optimal thermal placement and loss estimation for power electronic modules,'' \emph{IEEE Trans. Comp., Packag. and Manufacturing Tech.}, vol.~8, no.~2, pp. 236--243, 2018.

\bibitem{survey}
K.~Cao, J.~Zhou, T.~Wei, M.~Chen, S.~Hu, and K.~Li, ``A survey of optimization techniques for thermal-aware {3D} processors,'' \emph{Journal of Systems Architecture}, vol.~97, pp. 397--415, 2019.

\bibitem{survey2}
H.~Sultan, A.~Chauhan, and S.~R. Sarangi, ``A survey of chip-level thermal simulators,'' \emph{ACM Comput. Surv.}, vol.~52, no.~2, pp. 1--35, 2019.

\bibitem{3dic_thermal_reliability}
F.~Tavakkoli, S.~Ebrahimi, S.~Wang, and K.~Vafai, ``Analysis of critical thermal issues in 3d integrated circuits,'' \emph{International Journal of Heat and Mass Transfer}, vol.~97, pp. 337--352, 2016.

\bibitem{liu2014compact}
Z.~Liu, S.~Swarup, S.~X.-D. Tan, H.-B. Chen, and H.~Wang, ``Compact lateral thermal resistance model of {TSVs} for fast finite-difference based thermal analysis of {3-D} stacked {ICs},'' \emph{IEEE Transactions on Computer-Aided Design of Integrated Circuits and Systems}, vol.~33, no.~10, pp. 1490--1502, 2014.

\bibitem{li2004efficient}
P.~Li, L.~T. Pileggi, M.~Asheghi, and R.~Chandra, ``Efficient full-chip thermal modeling and analysis,'' in \emph{IEEE/ACM International Conference on Computer Aided Design, 2004. ICCAD-2004.}\hskip 1em plus 0.5em minus 0.4em\relax IEEE, 2004, pp. 319--326.

\bibitem{wang2004spice}
T.-Y. Wang and C.~C.-P. Chen, ``{SPICE-compatible} thermal simulation with lumped circuit modeling for thermal reliability analysis based on modeling order reduction,'' in \emph{International Symposium on Signals, Circuits and Systems}, 2004, pp. 357--362.

\bibitem{xie2013system}
J.~Xie and M.~Swaminathan, ``System-level thermal modeling using nonconformal domain decomposition and model-order reduction,'' \emph{IEEE Trans. CPMT}, vol.~4, no.~1, pp. 66--76, 2013.

\bibitem{dnn-thermal}
J.~Wen, S.~Pan, N.~Chang, W.-T. Chuang, W.~Xia, D.~Zhu, A.~Kumar, E.-C. Yang, K.~Srinivasan, and Y.-S. Li, ``Dnn-based fast static on-chip thermal solver,'' in \emph{Semiconductor Thermal Measurement, Modeling \& Management Symposium}, 2020, pp. 65--75.

\bibitem{ml-based-3d}
A.~Kumar, N.~Chang, D.~Geb, H.~He, S.~Pan, J.~Wen, S.~Asgari, M.~Abarham, and C.~Ortiz, ``{ML}-based fast on-chip transient thermal simulation for heterogeneous 2.5 {D}/{3D} {IC} designs,'' in \emph{International Symposium on VLSI Design, Automation and Test}, 2022, pp. 1--8.

\bibitem{ml-thermal}
R.~Ranade, H.~He, J.~Pathak, N.~Chang, A.~Kumar, and J.~Wen, ``A thermal machine learning solver for chip simulation,'' in \emph{ACM/IEEE Workshop on Machine Learning for CAD}, 2022, pp. 111--117.

\bibitem{geb2022chip}
D.~Geb, S.~Asgari, A.~Kumar, J.~Wen, N.~Chang, S.~Pan, M.~Abarham, H.~He, and V.~Gandhi, ``On-chip transient hot spot detection with a multiscale rom in 3dic designs,'' in \emph{2022 IEEE 72nd Electronic Components and Technology Conference (ECTC)}.\hskip 1em plus 0.5em minus 0.4em\relax IEEE, 2022, pp. 221--232.

\bibitem{wangaro}
M.~Wang, Y.~Cheng, W.~Zeng, Z.~Lu, V.~F. Pavlidis, and W.~W. Xing, ``{ARO:} autoregressive operator learning for transferable and multi-fidelity {3D-IC} thermal analysis with active learning,'' \emph{Proceedings of the IEEE/ACM International Conference on Computer-Aided Design (ICCAD)}, 2024.

\bibitem{smith2023real}
M.~J. Smith, S.~Hwang, V.~C. Do~Nascimento, Q.~Qiu, C.-K. Koh, G.~Subbarayan, and D.~Jiao, ``Real-time precision prediction of 3-d package thermal maps via image-to-image translation,'' in \emph{2023 IEEE 32nd Conference on Electrical Performance of Electronic Packaging and Systems (EPEPS)}.\hskip 1em plus 0.5em minus 0.4em\relax IEEE, 2023, pp. 1--3.

\bibitem{unsupervised-auto-encoder}
H.~He and J.~Pathak, ``An unsupervised learning approach to solving heat equations on chip based on auto encoder and image gradient,'' \emph{arXiv preprint arXiv:2007.09684}, 2020.

\bibitem{sanchis2022towards}
H.~Sanchis-Alepuz and M.~Stipsitz, ``Towards real time thermal simulations for design optimization using graph neural networks,'' in \emph{2022 IEEE Design Methodologies Conference (DMC)}.\hskip 1em plus 0.5em minus 0.4em\relax IEEE, 2022, pp. 1--6.

\bibitem{lu2024fast}
Z.~Lu, Y.~Zhou, Y.~Zhang, X.~Hu, Q.~Zhao, and X.~Hu, ``A fast general thermal simulation model based on multi-branch physics-informed deep operator neural network,'' \emph{Physics of Fluids}, vol.~36, no.~3, 2024.

\bibitem{ffn}
M.~Tancik, P.~Srinivasan, B.~Mildenhall, S.~Fridovich-Keil, N.~Raghavan, U.~Singhal, R.~Ramamoorthi, J.~Barron, and R.~Ng, ``Fourier features let networks learn high frequency functions in low dimensional domains,'' \emph{Advances in Neural Information Processing Systems}, vol.~33, pp. 7537--7547, 2020.

\bibitem{wang2021eigenvector}
S.~Wang, H.~Wang, and P.~Perdikaris, ``On the eigenvector bias of fourier feature networks: From regression to solving multi-scale pdes with physics-informed neural networks,'' \emph{Computer Methods in Applied Mechanics and Engineering}, vol. 384, p. 113938, 2021.

\bibitem{liu2024kan}
Z.~Liu, Y.~Wang, S.~Vaidya, F.~Ruehle, J.~Halverson, M.~Solja{\v{c}}i{\'c}, T.~Y. Hou, and M.~Tegmark, ``Kan: Kolmogorov-arnold networks,'' \emph{arXiv preprint arXiv:2404.19756}, 2024.

\bibitem{ss2024chebyshev}
S.~SS, K.~AR, A.~KP \emph{et~al.}, ``Chebyshev polynomial-based kolmogorov-arnold networks: An efficient architecture for nonlinear function approximation,'' \emph{arXiv preprint arXiv:2405.07200}, 2024.

\bibitem{yu2024separable}
X.~Yu, S.~Hooten, Z.~Liu, Y.~Zhao, M.~Fiorentino, T.~V. Vaerenbergh, and Z.~Zhang, ``Separable operator networks,'' \emph{Transactions on machine learning research}, 2024.

\bibitem{khan2015vector}
K.~A. Khan and P.~I. Barton, ``A vector forward mode of automatic differentiation for generalized derivative evaluation,'' \emph{Optimization Methods and Software}, vol.~30, no.~6, pp. 1185--1212, 2015.

\bibitem{saad1986gmres}
Y.~Saad and M.~H. Schultz, ``Gmres: A generalized minimal residual algorithm for solving nonsymmetric linear systems,'' \emph{SIAM Journal on scientific and statistical computing}, vol.~7, no.~3, pp. 856--869, 1986.

\bibitem{deeponet}
L.~Lu, P.~Jin, G.~Pang, Z.~Zhang, and G.~E. Karniadakis, ``Learning nonlinear operators via deeponet based on the universal approximation theorem of operators,'' \emph{Nature Machine Intelligence}, vol.~3, no.~3, pp. 218--229, 2021.

\bibitem{jin2022mionet}
P.~Jin, S.~Meng, and L.~Lu, ``Mionet: Learning multiple-input operators via tensor product,'' \emph{SIAM Journal on Scientific Computing}, vol.~44, no.~6, pp. A3490--A3514, 2022.

\bibitem{raissi2019physics}
M.~Raissi, P.~Perdikaris, and G.~E. Karniadakis, ``Physics-informed neural networks: A deep learning framework for solving forward and inverse problems involving nonlinear partial differential equations,'' \emph{Journal of Computational physics}, vol. 378, pp. 686--707, 2019.

\bibitem{wang2021learning}
S.~Wang, H.~Wang, and P.~Perdikaris, ``Learning the solution operator of parametric partial differential equations with physics-informed deeponets,'' \emph{Science advances}, vol.~7, no.~40, p. eabi8605, 2021.

\bibitem{ronen2019convergence}
B.~Ronen, D.~Jacobs, Y.~Kasten, and S.~Kritchman, ``The convergence rate of neural networks for learned functions of different frequencies,'' \emph{Advances in Neural Information Processing Systems}, vol.~32, 2019.

\bibitem{baydin2018automatic}
A.~G. Baydin, B.~A. Pearlmutter, A.~A. Radul, and J.~M. Siskind, ``Automatic differentiation in machine learning: a survey,'' \emph{Journal of machine learning research}, vol.~18, no. 153, pp. 1--43, 2018.

\bibitem{schmidt2021kolmogorov}
J.~Schmidt-Hieber, ``The {Kolmogorov--Arnold} representation theorem revisited,'' \emph{Neural networks}, vol. 137, pp. 119--126, 2021.

\bibitem{jacot2018neural}
A.~Jacot, F.~Gabriel, and C.~Hongler, ``Neural tangent kernel: Convergence and generalization in neural networks,'' \emph{Advances in neural information processing systems}, vol.~31, 2018.

\bibitem{lee2019wide}
J.~Lee, L.~Xiao, S.~Schoenholz, Y.~Bahri, R.~Novak, J.~Sohl-Dickstein, and J.~Pennington, ``Wide neural networks of any depth evolve as linear models under gradient descent,'' \emph{Advances in neural information processing systems}, vol.~32, 2019.

\bibitem{van1987simulated}
P.~J. Van~Laarhoven, E.~H. Aarts, P.~J. van Laarhoven, and E.~H. Aarts, \emph{Simulated annealing}.\hskip 1em plus 0.5em minus 0.4em\relax Springer, 1987.

\bibitem{nishino2017cupy}
R.~Nishino and S.~H.~C. Loomis, ``Cupy: A numpy-compatible library for nvidia gpu calculations,'' \emph{31st confernce on neural information processing systems}, vol. 151, no.~7, 2017.

\bibitem{akiba2019optuna}
T.~Akiba, S.~Sano, T.~Yanase, T.~Ohta, and M.~Koyama, ``Optuna: A next-generation hyperparameter optimization framework,'' in \emph{Proceedings of the 25th ACM SIGKDD international conference on knowledge discovery \& data mining}, 2019, pp. 2623--2631.

\end{thebibliography}

\appendix
\subsection{Visualization of All Test Cases for 2D Surface Power Map}\label{appendix:2d_all_cases}
\added{Figure \ref{fig:appendix_2d_all_cases} shows the complete set of 10 test cases used to compute the average MAPE values reported in Table \ref{tab:2d_comparison}. The error maps reveal that improvements are minimal for simple cases (columns 1-3) but become significant for complex multi-hotspot patterns (columns 4-10), with DeepOHeat-v1 particularly excelling at maintaining accuracy near thermal boundaries.}

\subsection{Detailed Derivation of Training Dynamics and Error Evolution}
\label{app:training_dynamics}
The target function \( f^*(y) \) on \( y \in [-1,1] \) is assumed to have a Chebyshev series expansion:
\begin{equation}
    f^*(y) = \sum_{k=0}^{\infty} b_k C_k(y),
\end{equation}
where \( \{C_k(y)\} \) are Chebyshev polynomials. We train a single-layer KAN model:
\begin{equation}
    f(y) = \sum_{k=0}^K a_k C_k(y),
\end{equation}
by minimizing the weighted \( L^2 \) loss:
\begin{equation}
    \mathcal{L} = \frac{1}{2}\int_{-1}^1 \left(f(y) - f^*(y)\right)^2 \frac{dy}{\sqrt{1-y^2}}.
\end{equation}
The gradient of \( \mathcal{L} \) with respect to parameter \( a_k \) is:
\begin{equation}
    \frac{\partial \mathcal{L}}{\partial a_k} =  \int_{-1}^1 \left(f(y) - f^*(y)\right) C_k(y) \frac{dy}{\sqrt{1-y^2}}.
\end{equation}
Under continuous-time gradient descent (gradient flow), parameters evolve as:
\begin{equation}
\begin{aligned}
    \frac{da_k(t)}{dt} &= -\eta \frac{\partial \mathcal{L}}{\partial a_k} \\&= -\eta \int_{-1}^1 \left(f(y,t) - f^*(y)\right) C_k(y) \frac{dy}{\sqrt{1-y^2}},
\end{aligned}
\end{equation}
where \( \eta > 0 \) is the learning rate. The time derivative of the network output \( f(y,t) = \sum_{k=0}^K a_k(t) C_k(y) \) becomes:
\begin{align}
\label{grad_flow}
    &\frac{\partial f(y,t)}{\partial t} = \sum_{k=0}^K \frac{da_k(t)}{dt} C_k(y) \nonumber \\
    &= -\eta \int_{-1}^1 \left[ \sum_{k=0}^K C_k(y)C_k(y') \right] \left(f(y',t)-f^*(y')\right)\frac{dy'}{\sqrt{1-y'^2}}.
\end{align}
This defines the Neural Tangent Kernel (NTK):
\begin{equation}
    \Theta_{\text{KAN}}(y,y') = \sum_{k=0}^K C_k(y)C_k(y').
\end{equation}

Define the error \( e(y,t) = f(y,t) - f^*(y) \). Projecting \( e(y,t) \) onto the Chebyshev basis:
\begin{equation}
    e(y,t) = \sum_{k=0}^K e_k(t) C_k(y),
\end{equation}
where the coefficients \( e_k(t) \) are computed via:
\begin{equation}
    e_k(t) = \frac{1}{\kappa_k} \int_{-1}^1 e(y,t) C_k(y) \frac{dy}{\sqrt{1-y^2}},
\end{equation}
and the normalization constants \( \kappa_k \) arise from Chebyshev orthogonality:
\begin{equation}
    \int_{-1}^1 C_i(y)C_j(y)\frac{dy}{\sqrt{1-y^2}} = \kappa_i \delta_{ij},
\end{equation}
where $\kappa_0 = \pi$, $\kappa_k = \frac{\pi}{2} \ (k \geq 1)$

Substitute \( e(y,t) = \sum_{k=0}^K e_k(t)C_k(y) \) into the gradient flow equation \eqref{grad_flow}:
\begin{align}
    \frac{\partial e(y,t)}{\partial t} &= -\eta \int_{-1}^1 \Theta_{\text{KAN}}(y,y') e(y',t) \frac{dy'}{\sqrt{1-y'^2}} \nonumber \\
    &= -\eta \sum_{k=0}^K \kappa_k e_k(t) C_k(y).
\end{align}
Equating coefficients for each \( C_k(y) \), we obtain decoupled ODEs:
\begin{equation}
    \frac{de_k(t)}{dt} = -\eta \kappa_k e_k(t),
\end{equation}
with solutions:
\begin{equation}
    e_k(t) = e_k(0) e^{-\eta \kappa_k t}.
\end{equation}
The total error evolution is therefore:
\begin{equation}
    e(y,t) = \sum_{k=0}^K e_k(0) e^{-\eta \kappa_k t} C_k(y).
\end{equation}
This contrasts with MLPs, where NTK eigenvalues typically decay exponentially with frequency \cite{jacot2018neural}, causing high-frequency modes to converge slower. KANs’ uniform spectral convergence enables simultaneous learning of multiscale features.

\subsection{Thermal Optimization Using Simulated Annealing}
\label{app:simulated_annealing}
\begin{algorithm}
\caption{Simulated Annealing for Thermal Optimization}
\label{alg:simulated_annealing}
\begin{algorithmic}[1]
\Require Initial design $\boldsymbol{u}_{ini}$, initial SA temperature $T_0$, cooling rate $\alpha$, iterations $N$, base neighborhood size $w_{base}$
\Ensure Optimized design $\boldsymbol{u}^*$
\State $\boldsymbol{u}_{\text{curr}} \gets \boldsymbol{u}_{ini}$
\State $T \gets T_0$
\State $\boldsymbol{u}^* \gets \boldsymbol{u}_{\text{curr}}$
\State $f^* \gets \text{EvaluateObjective}(\boldsymbol{u}^*)$
\State $f_{\text{curr}} \gets f^*$
\For{$i = 1$ to $N$}
    \State // Generate neighbor with adaptive neighborhood size
    \State $\text{temp\_ratio} \gets T/T_0$
    \State $\boldsymbol{u}_{\text{new}} \gets \boldsymbol{u}_{\text{curr}}$
    \For{each parameter $p$ in $\boldsymbol{u}_{\text{curr}}$}
        \State $\text{width} \gets \max(1, \lfloor w_{base} \cdot \text{temp\_ratio} \rfloor)$
        \State $\text{low} \gets \max(\boldsymbol{u}_{\text{curr}}[p] - \text{width}, \text{param\_min}[p])$
        \State $\text{high} \gets \min(\boldsymbol{u}_{\text{curr}}[p] + \text{width}, \text{param\_max}[p])$
        \State $\boldsymbol{u}_{\text{new}}[p] \gets \text{RandomInteger}(\text{low}, \text{high})$
    \EndFor
    \State // Evaluate objective function
    \State $f_{\text{new}} \gets \text{EvaluateObjective}(\boldsymbol{u}_{\text{new}})$
    \State // Accept or reject based on Metropolis criterion
    \If{$f_{\text{new}} \leq f_{\text{curr}}$ \textbf{or} $\exp((f_{\text{curr}} - f_{\text{new}})/T) > \text{random}(0,1)$}
        \State $\boldsymbol{u}_{\text{curr}} \gets \boldsymbol{u}_{\text{new}}$
        \State $f_{\text{curr}} \gets f_{\text{new}}$
        \If{$f_{\text{curr}} < f^*$}
            \State $\boldsymbol{u}^* \gets \boldsymbol{u}_{\text{curr}}$
            \State $f^* \gets f_{\text{curr}}$
        \EndIf
    \EndIf
    \State // Cool temperature
    \State $T \gets \alpha \cdot T$
\EndFor
\State \Return $\boldsymbol{u}^*$
\end{algorithmic}
\end{algorithm}

\end{document}